\documentclass[acmtog]{acmart}
\acmSubmissionID{157}

\usepackage{booktabs} 

\usepackage{bbm} 
\usepackage{amsfonts}
\usepackage{amsthm}
\usepackage{multirow}
\usepackage{enumitem}
\usepackage{framed}
\usepackage{mathtools}
\usepackage{breqn}
\usepackage{footmisc}
\usepackage{makecell}
\usepackage{footnote}
\usepackage[ruled]{algorithm2e} 

\SetAlFnt{\small}
\SetAlCapFnt{\small}
\SetAlCapNameFnt{\small}
\SetAlCapHSkip{0pt}

\newcommand{\ie}{\emph{i.e}.}
\newcommand{\eg}{\emph{e.g}.}

\acmJournal{TOG}

\setcopyright{acmcopyright}\acmJournal{TOG}
\acmYear{2022}\acmVolume{41}\acmNumber{6}\acmArticle{203}\acmMonth{12}
\acmDOI{10.1145/3550454.3555437}

\begin{document}
\title{VToonify: Controllable High-Resolution Portrait Video Style Transfer}

\author{Shuai Yang}
\orcid{0000-0002-5576-8629}
\affiliation{
 \institution{S-Lab, Nanyang Technological University}
 \country{Singapore}
}
\email{shuai.yang@ntu.edu.sg}

\author{Liming Jiang}
\orcid{0000-0001-8109-5598}
\affiliation{
 \institution{S-Lab, Nanyang Technological University}
 \country{Singapore}}
\email{liming002@ntu.edu.sg}

\author{Ziwei Liu}
\orcid{0000-0002-4220-5958}
\affiliation{
 \institution{S-Lab, Nanyang Technological University}
 \country{Singapore}
}
\email{ziwei.liu@ntu.edu.sg}

\author{Chen Change Loy}
\authornote{Corresponding Author}
\orcid{0000-0001-5345-1591}
\affiliation{
 \institution{S-Lab, Nanyang Technological University}
 \country{Singapore}}
\email{ccloy@ntu.edu.sg}

\begin{abstract}
Generating high-quality artistic portrait videos is an important and desirable task in computer graphics and vision. Although a series of successful portrait image toonification models built upon the powerful StyleGAN have been proposed, these image-oriented methods have obvious limitations when applied to videos, such as the fixed frame size, the requirement of face alignment, missing non-facial details and temporal inconsistency.
In this work, we investigate the challenging controllable high-resolution portrait video style transfer by introducing a novel \textbf{VToonify} framework. Specifically, VToonify leverages the mid- and high-resolution layers of StyleGAN to render high-quality artistic portraits based on the multi-scale content features extracted by an encoder to better preserve the frame details. The resulting fully convolutional architecture accepts non-aligned faces in videos of variable size as input, contributing to complete face regions with natural motions in the output. Our framework is compatible with existing StyleGAN-based image toonification models to extend them to video toonification, and inherits appealing features of these models for flexible style control on color and intensity. This work presents two instantiations of VToonify built upon Toonify and DualStyleGAN for collection-based and exemplar-based portrait video style transfer, respectively.
Extensive experimental results demonstrate the effectiveness of our proposed VToonify framework over existing methods in generating high-quality and temporally-coherent artistic portrait videos with flexible style controls. Code and pretrained models are available at our project page:
\url{www.mmlab-ntu.com/project/vtoonify/}.
\end{abstract}

%
%

\ccsdesc[500]{Computing methodologies~Computational photography}
\ccsdesc[300]{Computing methodologies~Computer vision}

%
%

\keywords{Face toonification, exemplar-based, model distillation, StyleGAN}

\begin{teaserfigure}
  \includegraphics[width=\textwidth]{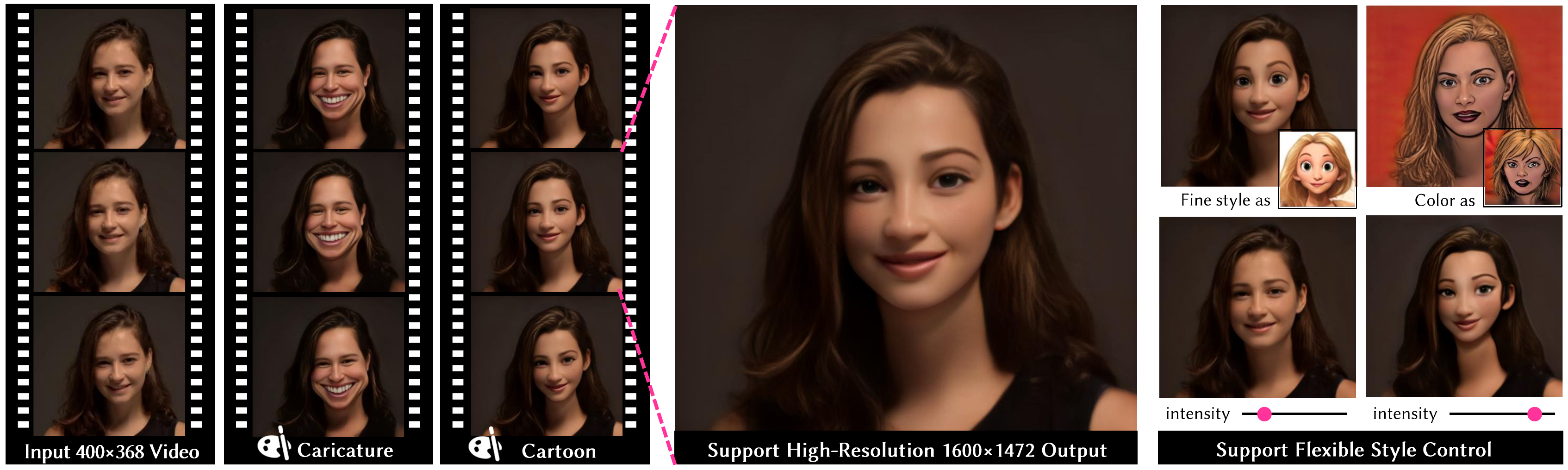}
  \caption{\textbf{Controllable High-Resolution Portrait Video Style Transfer.} We propose a novel VToonify framework that can synthesize high-resolution stylized videos from a low-resolution input. VToonify precisely pastiches the facial structure style of style collections of Caricature and Cartoon. Within each style collection, fine-level style transfer in terms of both structure style and color style is supported by specifying a reference style image. Besides style types, our framework further supports the adjustment of style degrees to flexibly choose to which degree the original facial features are preserved. Input video: \copyright Pexels Ketut Subiyanto.} 
  \label{fig:teaser}
\end{teaserfigure}

\maketitle

\section{Introduction}

Artistic portraits are ubiquitous in our daily life as well as creative industries in forms of arts, social media avatars, movies, entertainment advertising, \textit{etc}.
With the advent of deep learning technology, one can now render high-quality artistic portraits from real face images through automatic portrait style transfer.
There are a number of successful approaches designed for image-based style transfer~\cite{gatys2016image, selim2016painting,pinkney2020resolution,song2021agilegan,alaluf2021restyle,jang2021stylecarigan,yang2022Pastiche}, with many of them easily accessible to novice users in the form of mobile applications.
Over the last few years, video content has quickly become a staple of our social media feeds.
The popularity of social media and ephemeral videos arouse increasing demands for creative editing of videos, such as portrait video style transfer, to create effective and engaging videos.
Existing image-oriented approaches have several drawbacks when applied to videos, limiting their applications in automatic portrait video stylization.

\begin{figure*}[t]
\centering
\includegraphics[width=0.97\linewidth]{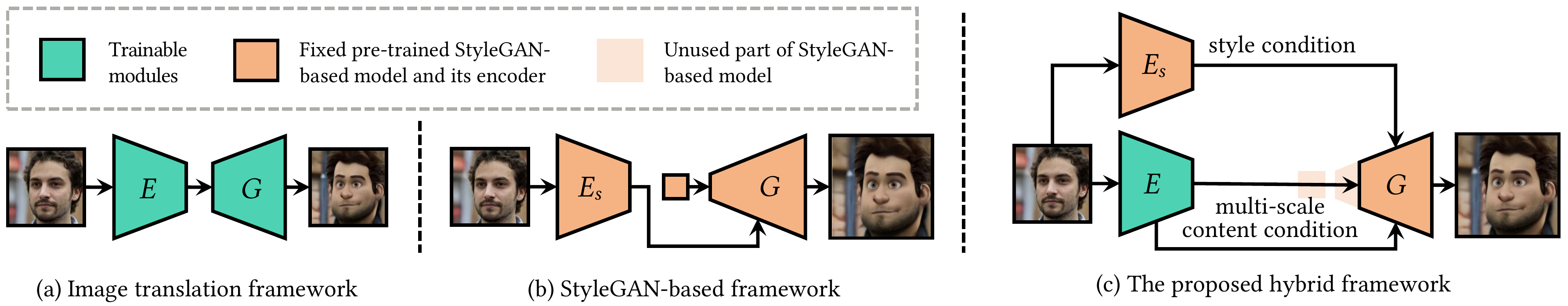}
\caption{\textbf{Overview of the proposed hybrid framework.} (a) Image translation framework uses fully convolutional networks to support variable input size. However, training from scratch renders it difficult for high-resolution and controllable style transfer. (b) The StyleGAN-based framework leverages the pre-trained model for high-resolution and controllable style transfer, but is limited to fixed image size and detail losses. (c) Our hybrid framework combines the above two ideas, supporting high-resolution and controllable style transfer on videos of various sizes.}
\label{fig:main_idea}
\end{figure*}

The ability to generate high-quality 1024$\times$1024 faces with flexible style control makes StyleGAN a popular backbone for building a portrait image style transfer model. Such a StyleGAN-based framework (also called image toonification) mainly encodes a real face into the StyleGAN latent space, and applies the resulting style code to another StyleGAN fine-tuned on the artistic portrait dataset to obtain its stylized version~\cite{pinkney2020resolution}.
A key problem with this framework is that StyleGAN produces images with aligned faces and under a fixed size, which does not favor dynamic faces in real videos like Fig.~\ref{fig:teaser}. Cropping and aligning faces in the video often results in an incomplete face and unnatural motions. We refer to this problem as the `fixed-crop limitation' of StyleGAN.
Although StyleGAN3~\cite{karras2021alias} has been proposed for unaligned faces, it only supports a fixed image size. Moreover, compared to aligned faces, a recent study~\cite{alaluf2022times} finds that it becomes more difficult to encode unaligned faces.
An inaccurate face encoding is detrimental to portrait style transfer, causing problems such as identity change and missing items in the reconstructed frames and the stylized frames.

Different from the StyleGAN-based framework, image-to-image translation directly learns to map the input image from the real face domain to the artistic portrait domain. This problem is often tackled using a fully convolutional encoder-generator architecture, eschewing strict restrictions on the image size and the face positions during the test phase.
Due to the lack of paired training data, the framework typically builds upon the cycle consistency paradigm~\cite{Zhu2017Unpaired} to learn bi-directional mappings.
However, the complicated mappings constrain the models to a small image size of 256$\times$256.
There are also frameworks~\cite{viazovetskyi2020stylegan2} distilling StyleGAN for face attribute editing, where the StyleGAN is used to generate paired data for training. The strong paired supervision enables the framework to handle high-resolution images.
Nevertheless, the above method itself does not incorporate StyleGAN into its network design. Thus, StyleGAN's good properties of flexible style control are unfortunately not inherited. 
Recently, GLEAN~\cite{chan2020glean} introduces StyleGAN to the translation network design to effectively use the StyleGAN face priors but entails the accompanying `fixed-crop limitation'.  
Despite the limitation, this hybrid framework shows high performance and inspires our work. 

As analyzed above, an effective method needs to address the following challenges for portrait video style transfer:
\textbf{1)} The method needs to cope with unaligned faces and different video sizes to maintain natural motions.
Large video size, or wide angle of view, can capture more information and prevent the face from moving out of the frame.
\textbf{2)} Generating high-resolution videos is desired to match the widely used HD devices nowadays.
\textbf{3)} To build a practical user interaction system, flexible style control should be provided for users to adjust and select their preference.

To this end, we propose a new hybrid framework \textbf{VToonify} for video toonification.
We first analyze the translation equivariance in StyleGAN which constitutes our key solution to overcome the fixed-crop limitation.
As shown in Fig.~\ref{fig:main_idea}(c), VToonify combines the merits of the StyleGAN-based framework~\cite{pinkney2020resolution} and the image translation framework~\cite{viazovetskyi2020stylegan2}, achieving controllable high-resolution portrait video style transfer.
We adopt the StyleGAN architecture following~\cite{pinkney2020resolution} for high-resolution style transfer, but adapt StyleGAN by removing its fixed-sized input feature and low-resolution layers to construct a novel fully convolutional encoder-generator architecture akin to that in the image translation framework, which supports different video sizes.
Apart from the original high-level style code, we train an encoder to extract multi-scale content features of the input frame as the additional content condition to the generator, so that the key visual information of the frame can be better preserved during style transfer.
We follow~\cite{viazovetskyi2020stylegan2,chen2019animegan} to distill StyleGAN on its synthesized paired data. We further propose a flicker suppression loss based on the simulation of camera motion over a single synthetic data to eliminate flickers. Therefore, VToonify is able to learn a fast and coherent video translation without real data, complicated video synthesis or explicit optical flow computation.
Different from the standard image translation framework in \cite{viazovetskyi2020stylegan2,chen2019animegan}, VToonify incorporates the StyleGAN model into the generator to distill both data and model.
Therefore, VToonify inherits the style adjustment flexibility of StyleGAN. By reusing StyleGAN as the generator, we only need to train the encoder, greatly reducing both the training time and training difficulty. 

With the novel formulation above, this paper presents two variants of VToonify built upon two representative StyleGAN backbones, Toonify~\cite{pinkney2020resolution} and DualStyleGAN~\cite{yang2022Pastiche}, for collection-based and exemplar-based portrait video toonification, respectively. The former stylizes faces based on the overall style of the dataset, while the later uses a single image in the dataset to specify finer-level style as shown on the top right of Fig.~\ref{fig:teaser}.
By adapting DualStyleGAN's style control modules~\cite{yang2022Pastiche} to adjust the features of our encoder and elaborately designing the data generation and training objectives, VToonify inherits DualStyleGAN's flexible style control and adjustment of style degrees, and further extends these features to videos (\eg, the top right of Fig.~\ref{fig:teaser}).
We show in the experiment that VToonify generates stylized frames not only as high-quality as the backbones but also better preserving the details of the input frame.
To summarize, our main contributions are as follows:
\begin{itemize}
  \item We analyze the fixed-crop limitation of StyleGAN and suggest a corresponding solution based on the translation equivariance in StyleGAN.
  \item We propose a novel fully convolutional VToonify framework for controllable high-resolution portrait video style transfer, supporting unaligned faces and various video sizes.
  \item We build VToonify upon Toonify and DualStyleGAN backbones and distill the backbones in terms of both data and model, to realize collection-based and exemplar-based portrait video style transfer.
  \item We design a principled data-friendly training scheme and propose an optical-flow-free flicker suppression loss for temporal consistency, which is efficient and effective for training a video style transfer model.
\end{itemize}

\section{Related Work}

\subsection{Face Image Style Transfer}

\paragraph{Image-to-Image Translation}
To learn the structure style, image translation framework is adopted for face style transfer.
Pix2pix~\cite{Isola2017Image} first proposes the supervised image translation framework to map images from a source domain to a target domain. This framework requires paired data for training, which is not easy to collect for face style transfer.
To solve this problem, some methods like AnimeGAN2~\cite{chen2019animegan} distill existing style transfer models~\cite{pinkney2020resolution} by generating synthetic face-cartoon pairs for training.
Meanwhile, CariGAN~\cite{cao2018carigans} solves a weakly supervised task where the paired face and caricature share the same identity but without pixel-level correspondences. 

To support unpaired data, cycle consistency is proposed in CycleGAN~\cite{Zhu2017Unpaired} for unsupervised image translation.
The following face-to-cartoon translation methods improve CycleGAN by incorporating attention to the domain-specific features~\cite{kim2019u}, data augmentation to strengthen domain invariance~\cite{chong2021gans}, and matching multi-scale statistical features to stabilize training~\cite{shao2021spatchgan}. Cycle consistency sometimes can be restrictive for imbalanced domains, \eg, transforming abstracted anime face back to real faces. To solve this problem, CouncilGAN~\cite{nizan2020breaking} leverages collaboration between multiple generators instead of cycle consistency while AniGAN~\cite{li2021anigan} constrains translations with multi-scale style losses.
One main drawback of the unpaired image translation framework is that learning from scratch on complicated translations makes this framework limited to low-resolution images.

By comparison, besides distilling data, we further incorporate the StyleGAN model into our network design, thus supporting both high-resolution video generation and flexible style control.

\paragraph{StyleGAN-Based Toonification}
StyleGAN~\cite{karras2019style,karras2020analyzing} generates photo-realistic face images and provides hierarchical style control, serving as a powerful backbone for face stylization.
Toonify~\cite{pinkney2020resolution} fine-tunes a pre-trained StyleGAN on a cartoon dataset and combines the shallow layers of the fine-tuned model with the deep layers of the original model to generate faces in cartoon structures while maintaining the realistic facial color and textures.
The pSp method~\cite{richardson2020encoding} accelerates Toonify by training an encoder to project real face images into the closest cartoon faces in the fine-tuned latent space.
AgileGAN~\cite{song2021agilegan} and ReStyle~\cite{alaluf2021restyle} improve pSp by a variational encoder and an iterative refinement mechanism, respectively.
StyleCariGAN~\cite{jang2021stylecarigan} combines StyleGAN with image translation by explicitly learning structure transfer with cycle translations.
DualStyleGAN~\cite{yang2022Pastiche} extends StyleGAN with an extrinsic style path to accept the condition from a style image for exemplar-based style transfer. 
StyleGAN-NADA~\cite{gal2022stylegan} proposes to shift StyleGAN to new artistic domains guided by CLIP~\cite{radford2021learning} without using any real cartoon dataset, realizing text-driven toonification.

Although high-quality results are shown, the above methods assume a face is well aligned, which is unrealistic for videos.
By comparison, VToonify supports end-to-end unaligned face stylization in videos of variable size.

\subsection{StyleGAN Inversion}

StyleGAN inversion aims to project real face images into the latent space of StyleGAN for editing.
Image2StyleGAN~\cite{abdal2019image2stylegan} proposes to project real images into the $\mathcal{W}+$ space through optimization. PSp~\cite{richardson2020encoding} trains an encoder to accelerate the projection.
E4e~\cite{tov2021designing} predicts the latent codes in $\mathcal{W}+$  that reside close to the $\mathcal{W}$ space to improve the editability. Instead of searching the optimal latent code, PTI~\cite{roich2021pivotal} fine-tunes StyleGAN to approach the target image based on its predicted $\mathcal{W}$-space latent code. Based on this idea, HyperInverter~\cite{dinh2022hyperinverter} and HyperStyle~\cite{alaluf2022hyperstyle} propose to train a hyper network to directly predict the offsets of the StyleGAN parameters to simulate the fine-tuning in a more efficient way.
Recently, methods~\cite{wang2022high,zhu2021barbershop,parmar2022spatially} have been proposed to invert real images into the feature space ($\mathcal{F}$ space) of StyleGAN, which could better reconstruct the image details and is suitable for spatial editing. Our encoder also extracts content features in addition to the latent code as the condition to StyleGAN, which can be seen as a hybrid $\mathcal{F}$-$\mathcal{W}+$ space.
Different from the above methods, our method uses the $\mathcal{F}$ space to overcome StyleGAN's  fixed-crop limitation. We refer to~\cite{xia2022gan} for a comprehensive survey on StyleGAN inversion.

\subsection{Video Style Transfer}

\paragraph{Optical-Flow-Based Method}
Video style transfer pays additional attentions to the temporal consistency, which is usually achieved by optical flows.
\cite{ruder2016artistic} and \cite{gupta2017characterizing} apply Neural Style Transfer~\cite{gatys2016image} to videos by using the optical flow of the input video to warp the previous stylized frame to the current frame for frame initialization or extra conditions. \cite{huang2017real} and \cite{chen2017coherent} use a temporal consistency loss to constrain the coherence between corresponding pixels of consecutive stylized frames.
\cite{wang2020consistent} propose a compound regularization to better fit the nature of temporal variation with both optical flows and local jitters. 
The above approaches assume identical optical flows between the input frames and between the stylized frames, which does not hold for toonification where the faces are often deformed.
Unlike previous studies, we formulate an effective and simple-to-implement flicker suppression loss by simulating the motion of camera over a single frame.

\paragraph{Image-Animation-Based Method}
Another way of synthesizing artistic portrait videos is to apply image animation methods like First Order Motion~\cite{siarohin2019first} to animate an artistic portrait with the motion driven by the video.
Recently, DaGAN~\cite{hong2022depth} shows promising results by estimating the dense 3D geometry from the face videos to constrain the output to be more consistent with 3D face structures.
Such methods, however, only use the information of a single stylized frame, inevitably losing important details.
As we will show later, even the 3D geometry is considered, the motion of the facial textureless region is still hard to predict, leading to annoying flickers.

\paragraph{StyleGAN-Based Video Editing}
Besides image editing~\cite{harkonen2020ganspace,shen2020interpreting,shen2021closed,jiang2021talk}, StyleGAN has also gained attention in video editing.
\cite{fox2021stylevideogan} project video frames into the sequences of StyleGAN latent codes, and train a network to capture the temporal correlations from these low-dimensional codes.
\cite{yao2021latent} train a latent transformation network that disentangles the identity and facial attributes to edit the face with better identity preservation.
\cite{liu2022deepface} design a sketch branch combined with StyleGAN for sketch-based video editing.
To seamlessly stitch the cropped edited face and the background, STIT~\cite{tzaban2022stitch} proposes to tune StyleGAN to provide spatially-consistent transitions.
All the above methods require face alignment and cropping as pre-processing.
While StyleGAN3~\cite{karras2021alias} is proposed to support non-aligned faces, recent study~\cite{alaluf2022times} shows that such pre-processing is still needed for valid encoding in StyleGAN3, followed by projecting the results back to the original context. Results generated through such a complicated process are prone to artifacts.

By comparison, VToonify is an end-to-end framework without the above data pre-processing and post-processing.

\begin{figure}[t]
\centering
\includegraphics[width=\linewidth]{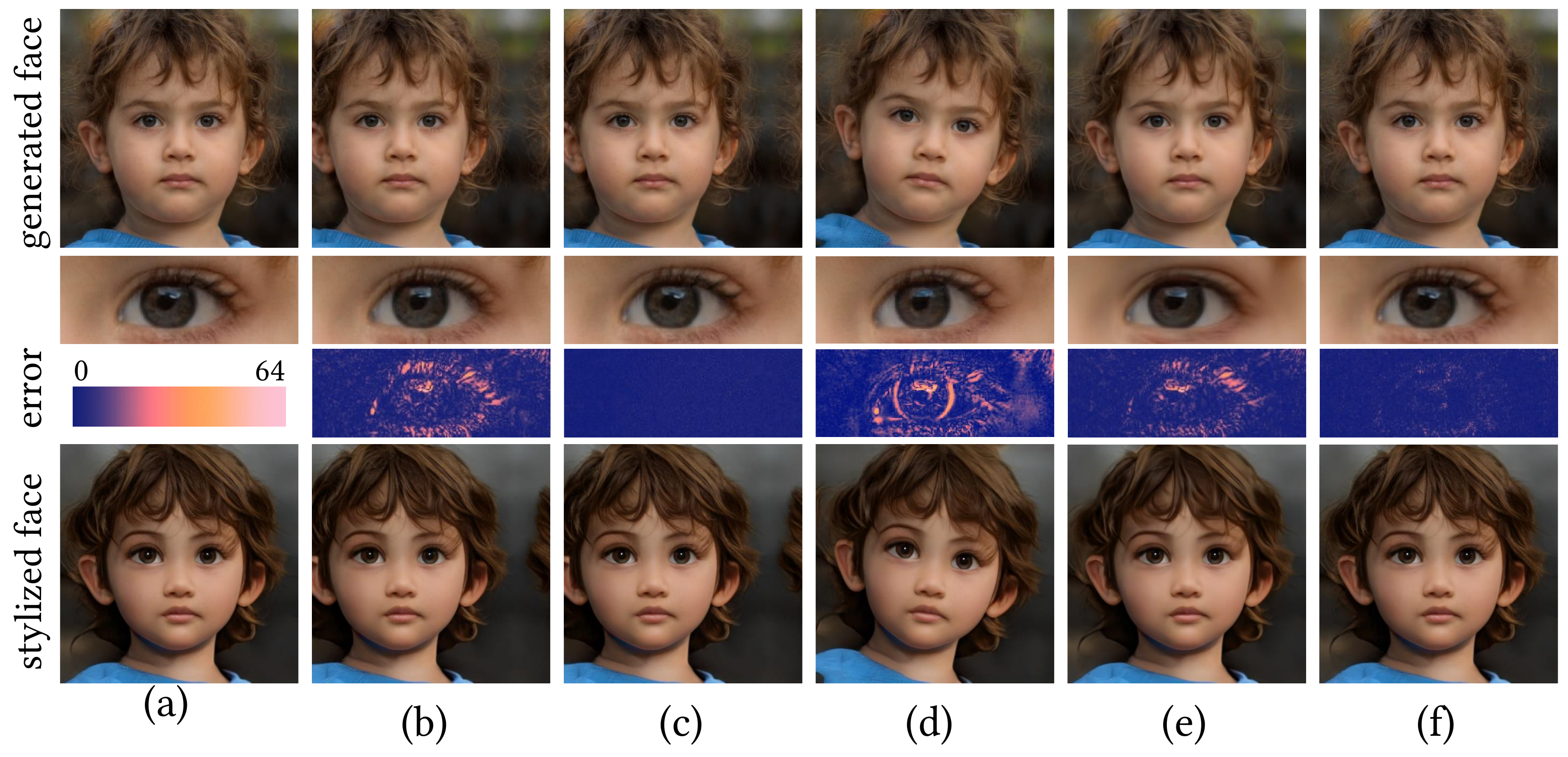}
\caption{\textbf{Analysis of the StyleGAN} in generating alias free unaligned faces. (a) Generated face by StyleGAN. (b) Face generated by translating the feature map of the StyleGAN 7-th layer by 128 pixels. Compared to (b), (c) additionally translates noise maps after the 7-th layer. (d)  Face generated by rotating the feature map of the StyleGAN 7-th layer 10 degrees. (e) Face generated by removing the noise inputs after the 7-th layer. (f) Face generated by removing the noise inputs after the 13-th layer. Each groups show the generated face, the enlarged (re-aligned) right eyes, the error maps with (a), and the toonified version by DualStyleGAN.}
\label{fig:analysis}
\end{figure}

\section{Analysis of StyleGAN-Based Toonification}
\label{sec:analysis}

Before we start introducing VToonify, we first briefly look into the problem of StyleGAN-based toonification when applied to video.
We will analyze the causes of the problem and present our corresponding solutions. Here we focus on StyleGAN/StyleGAN2~\cite{karras2019style,karras2020analyzing},
since most existing StyleGAN-based toonification models are built upon them.

\paragraph{Can a pre-trained StyleGAN generate unaligned faces?} The translation equivariance of convolutional networks naturally supports face translation.
As shown in Fig.~\ref{fig:analysis}(b), we first obtain the 32$\times$32 feature from the 7-th layer of StyleGAN, and apply translation to it with reflection padding. The translated feature is decoded by the last 11 layers of StyleGAN, which as expected is exactly the translated version of the original face.
We further try feature rotations and find that the output faces are rotated and look still plausible within a small rotation ranges in Fig.~\ref{fig:analysis}(d).
This implies that the last 11 layers of StyleGAN, which mainly render colors and textures for faces, is robust to geometric transformations and could be used for high-resolution unaligned face generation.

\paragraph{Can a pre-trained StyleGAN generate images under different sizes?} The fully convolutional architecture naturally supports different input sizes since the convolution operation is independent of input size. The reason why StyleGAN is limited to a fixed size is its fixed-sized input feature as shown in Fig.~\ref{fig:main_idea}(b). By removing this fixed input and adding another encoder for providing features of variable size, we can build a fully convolutional encoder-generator architecture as in neural style transfer~\cite{johnson2016perceptual} and image translation~\cite{Zhu2017Unpaired} for different input sizes. Moreover, the encoder directly extracts 32$\times$32 features from unaligned faces instead of artificially aligning faces and transforming their features.

\begin{figure*}[t]
\centering
\includegraphics[width=\linewidth]{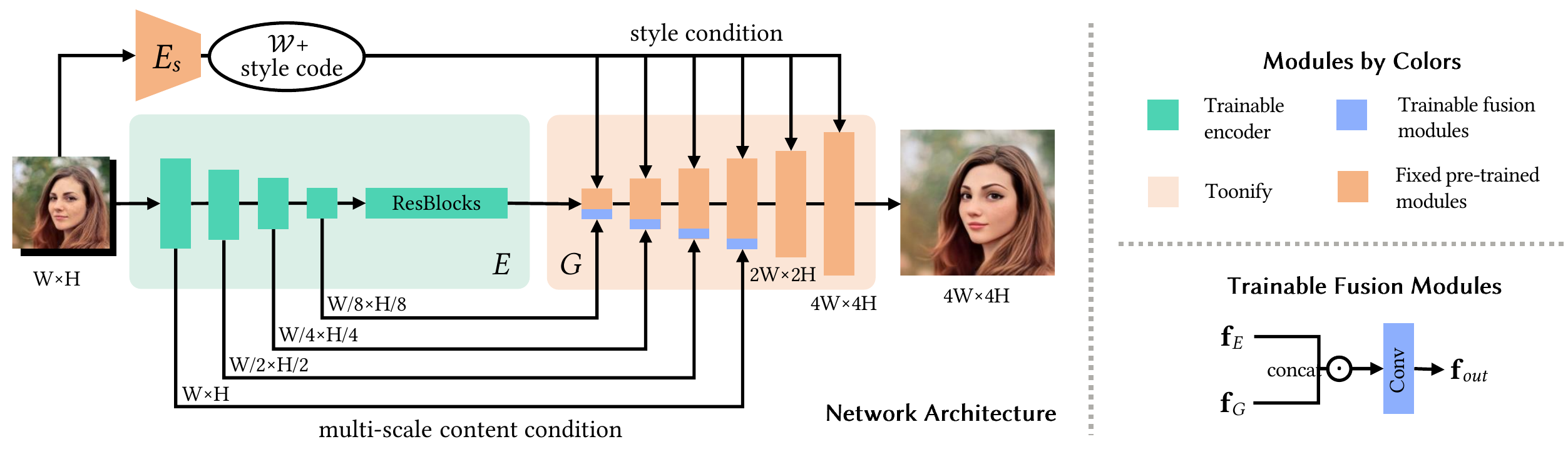}
\caption{\textbf{Framework of the proposed collection-based portrait video style transfer.} We use the highest 11 layers of the pre-trained Toonify as our generator $G$. In addition to the style code as condition, $G$ is also conditioned on the multi-scale content features from the encoder $E$. In the training phase, we use 256$\times$256 images as input, while during testing, our fully convolutional network accepts frames of various sizes. Input video: \copyright Pexels Andrea Piacquadio.}
\label{fig:framework1}
\end{figure*}

\paragraph{Where do flickers come from?} In our experiment on translating the 32$\times$32 feature, we notice flickers in the output. If we compare the translated face with the original one, we can observe the changes in textures as visualized by the error map in Fig.~\ref{fig:analysis}(b).
The cause is the additional noise inputs to enrich texture details. Random noises create random texture details while a fixed noise leads to
`texture sticking' artifacts~\cite{karras2021alias}. Removing the noise inputs can solve this problem but at cost of missing details as in Fig.~\ref{fig:analysis}(e).
The problem can also be solved by translating the noise maps in the last 11 layers along with the 32$\times$32 feature (Fig.~\ref{fig:analysis}(c)). However, the motion in real applications is more complex and the face deformation during style transfer makes the estimation of such motion in each layer impractical.
Our solution is to remove the noise inputs and to pass the multi-scale encoder features to the generator to supplement the texture details, as illustrated in Fig.~\ref{fig:main_idea}(c).
We further find that the noise inputs to the high-resolution layers have little effect as in Fig.~\ref{fig:analysis}(f), meaning high-resolution encoder features are not necessary. Thus, we can safely follow the common practices to use low-resolution videos as input to generate high-resolution videos, which improves efficiency and reduces the requirement for input resolution.

\section{Vtoonify for Video Toonification}

\subsection{Collection-Based Portrait Video Style Transfer}
\label{sec:collection}

We start from a simple case: collection-based portrait video style transfer.
The collection-based task transfers a holistic style of the style collection. 
We leverage the representative Toonify~\cite{pinkney2020resolution} as the backbone, which uses the original StyleGAN architecture and is solely conditioned on the style code.

\subsubsection{Network Architecture}

As shown in Fig.~\ref{fig:framework1}, the collection-based VToonify framework contains an encoder $E$ and a generator $G$ built upon Toonify.
$E$ accepts video frames and generates content features, which are then fed into $G$ to produce the final stylized portraits.
Unlike existing StyleGAN-based frameworks that use the whole StyleGAN architecture, we only use the highest $11$ layers of StyleGAN to build our $G$.
As analyzed in \cite{karras2019style}, the low-resolution and high-resolution layers of StyleGAN mainly capture structure-related styles and color/texture styles, respectively. Therefore, the main task of $G$ is to upsample the content features and render stylish colors and textures for them.

Following the common practices of StyleGAN-based framework, we incorporate a StyleGAN encoder $E_s$ to provide $G$ with its style condition. Specifically, $E_s$ embeds the input frame into the StyleGAN $\mathcal{W}+$ latent space with eighteen $512$-dimensional vectors, serving as the inputs to the eighteen layers of StyleGAN.
We use the correspondingly last 11 vectors as the style condition of $G$.

Besides the style condition, $E$  downsamples the input $W\times H$ frame and extract its $W/8\times H/8$ content feature with downsampling convolutional layers and ResBlocks~\cite{he2016deep}. The content feature replaces the original fixed-sized input features of StyleGAN in the $8$-th layer, so that variable frame size is allowed in our framework (as long as $W$ and $H$ are divisible by $8$).

As analyzed in Sec.~\ref{sec:analysis}, the application of noises enriches the texture details but creates flickers on videos. To solve this problem, we remove the noise input 
and pass the mid-layer features  of $E$ with rich low-level information to the corresponding resolution layers of $G$. Within each layer of $G$, we add a new fusion module to combine the generator feature $\mathbf{f}_G$ and the passed encoder feature $\mathbf{f}_E$. As shown in the bottom right of Fig.~\ref{fig:framework1}, $\mathbf{f}_G$ and $\mathbf{f}_E$ are concatenated channel-wisely as $\mathbf{f}_G\odot\mathbf{f}_E$ and then go through a convolutional layer to obtain the fused feature $\mathbf{f}_{out}$ as the input of the generator's next layer. The mid-layer features and the last-layer content feature form multi-scale content condition for $G$, effectively preserving the image details of the input frame compared to only using the style condition in previous StyleGAN-based approaches.

In summary, we remove the fixed-sized low-resolution layers of StyleGAN to support variable input size,
and replace the noise inputs with multi-scale content conditions to prevent flickers and preserve the input frame details.

\begin{figure}[t]
\centering
\includegraphics[width=\linewidth]{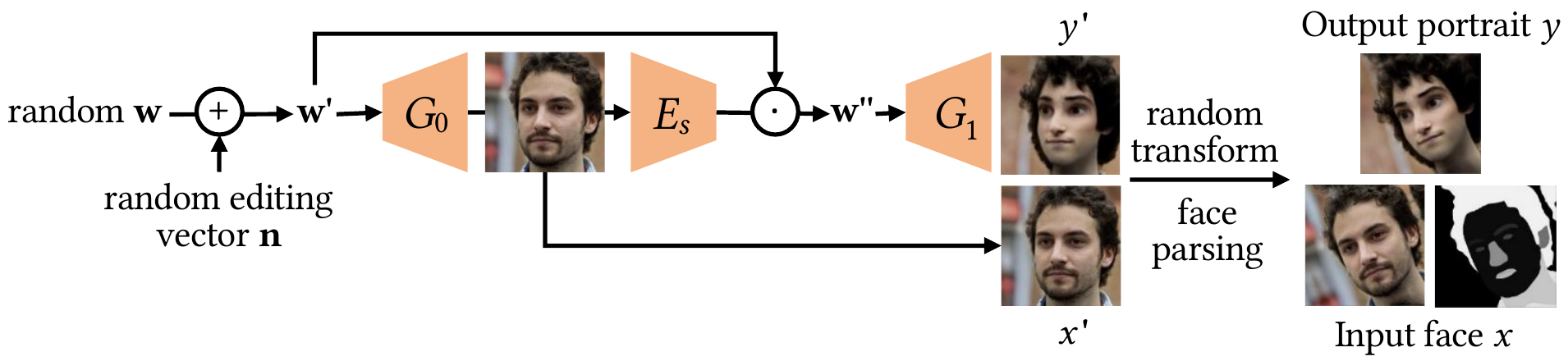}
\caption{\textbf{Pipeline of paired training data generation} for training collection-based portrait style transfer.}
\label{fig:data_generation}
\end{figure}

\subsubsection{Data Generation and Training Objectives}
\label{sec:data_generation1}

\paragraph{Data Generation} To distill Toonify for high-resolution toonification, we generate paired data based on it to train our model. The data generation process is intuitively illustrated in Fig.~\ref{fig:data_generation}.
Here, we use the original StyleGAN $G_0$ trained on real human faces and the fine-tuned StyleGAN $G_1$ in Toonify ($G$ is the highest 11 layers of $G_1$) to synthesize the 1024$\times$1024 face image $x$ and its corresponding 1024$\times$1024 ground truth stylized portrait $y$, respectively.

Specifically, we first sample random style codes $\mathbf{w}\in\mathcal{W}+$ 
to synthesize face images with $G_0$.
StyleGAN tends to generate smiling faces with eyes open and mouths half-opened, introducing significant bias into the data. To solve this problem, we leverage the editing vectors $\mathbf{n}$ obtained from \cite{zhu2021low} to explicitly edit these face attributes. The vectors $\mathbf{n}$ are randomly sampled from \{do nothing, remove smile, close eyes, close mouth, open mouth\} and added to $\mathbf{w}$ to obtain the edited style code $\mathbf{w}'$. The corresponding face with edited attribute is then synthesized as $x'=G_0(\mathbf{w}')$. To obtain its stylized counterpart, we further compute its embedded style code $E_s(x'_\downarrow)$ ($E_s$ requires 256$\times$256 inputs so we downsample $x'$ to $x'_\downarrow$ by 4$\times$) and generate a new style code $\mathbf{w}''$ by concatenating the first 7 vectors of $\mathbf{w}'$ and the last 11 vectors of $E_s(x'_\downarrow)$. And the stylized portrait is obtained by $y'=G_1(\mathbf{w}'')$. Note that in Toonify, stylized portraits are normally obtained as $G_1(\mathbf{w}')$ for synthetic face inputs or $G_1(E_s(x'_\downarrow))$ for real face inputs. By comparison, our $y'$ uses the structure style part of $\mathbf{w}'$ to maintain better identity of $x'$, and the color and texture style part of $E_s(x'_\downarrow)$ to ensure its consistency with the form of style condition in our framework (see Fig.~\ref{fig:framework1}).

Finally, to make our model learn to handle unaligned faces, we augment $x'$ and $y'$ with random geometric transformations~\cite{karras2020training} like flipping, scaling, translation and rotation. The resulting \{$x$, $y$\} together with the style code $\mathbf{w}''$ form the paired training data of our method. To help the network better identify the face regions of $x$ especially at the beginning of training, we obtain its face parsing map by applying BiSeNet~\cite{Yu-ECCV-BiSeNet-2018} to $x$, which serves as the additional semantic channels of $x$ in our framework (\ie, the black channels concatenated to the input frame in Fig.~\ref{fig:framework1}). For simplicity, we will use $x$ to denote the concatenation of the face image and its parsing map in the following.

\begin{figure}[t]
\centering
\includegraphics[width=\linewidth]{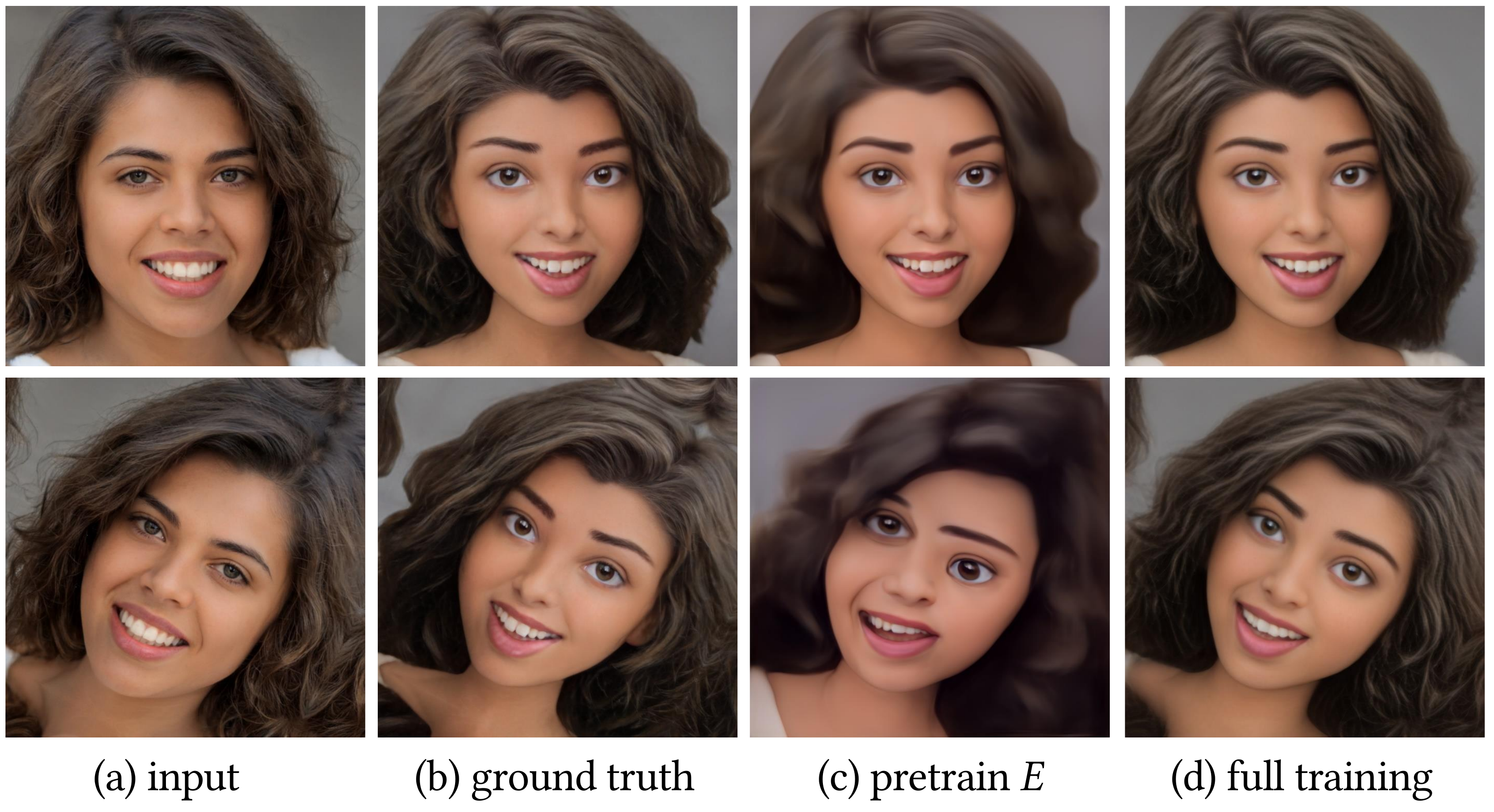}
\caption{\textbf{Training target and results of each stage.} (a) and (b) are the synthetic paired data for training VToonify. Bottom row applies geometric transformations. (c) Results of VToonify after pre-training $E$ and initializing the fusion modules. (d) Results of VToonify after full training.}
\label{fig:pretrain}
\end{figure}

\paragraph{Model Initialization} Different from the well-defined $G$, $E$ is randomly initialized. To stabilize the training, we first pre-train $E$. Since the last-layer feature of $E$ replaces the original 8-th-layer input feature of $G_1$, we would like these two features match. Let $\mathbf{f}^{(\text{last})}_{E}(x)$ denote the last-layer feature of $E$ given a frame $x$ and $\mathbf{f}^{(8)}_{G_1}(\mathbf{w})$ denote the 8-th-layer input feature of $G_1$ given a style code $\mathbf{w}$. We pre-train $E$ to minimize the objective:
\begin{equation}
\label{eq:pretrain_loss}
  \mathcal{L}_{\text{E}}=\|\mathbf{f}^{(\text{last})}_{E}(x'_\downarrow)-\mathbf{f}^{(8)}_{G_1}(\mathbf{w}'')\|_2.
\end{equation}
Besides, we initialize the fusion modules to map $\mathbf{f}_G\odot\mathbf{f}_E$  to $\mathbf{f}_G$.
This is realized by setting the half part of the convolution kernel that interacts with $\mathbf{f}_G$ to the identity matrix and the other half part that interacts with $\mathbf{f}_E$ to values close to 0.
With the above pre-trained encoder and initialization, our model exactly mimics the function of $G_1$ without noise inputs and could generate a rough stylized portrait as shown in Fig.~\ref{fig:pretrain}(c). Up to this point, our model is ready for the subsequent training to generate facial details and support unaligned faces as in Fig.~\ref{fig:pretrain}(d).

\paragraph{Training Objectives} We would like to stylize the low-resolution $x_\downarrow$ ($4\times$ downsampling of $x$) to approach the high-resolution ground truth $y$, leading to a reconstruction loss:
\begin{equation}
\label{eq:reconstruction_loss}
  \mathcal{L}_{\text{rec}}=\lambda_{\text{mse}}\|\widetilde{y}-y\|_2+\lambda_{\text{perc}}\mathcal{L}_{\text{perc}}(\widetilde{y},y),
\end{equation}
where $\widetilde{y}=G(E(x_\downarrow), \mathbf{w}'')$  is the stylization result given the content condition $E(x_\downarrow)$ and the style condition $\mathbf{w}''$. $\mathcal{L}_{\text{perc}}(\widetilde{y},y)$ is the perceptual loss~\cite{johnson2016perceptual} to minimize the semantic difference between $\widetilde{y}$ and $y$. Hyper-parameters $\lambda$s are used to balance different loss terms.
The realism of $\widetilde{y}$ is further reinforced through an adversarial training with a discriminator $D$,
\begin{equation}
  \mathcal{L}_{\text{adv}}=\mathbb{E}_{y}[\log D(y)]+\mathbb{E}_{x}[\log (1-D(\widetilde{y}))].
\end{equation}

\begin{figure}[t]
\centering
\includegraphics[width=\linewidth]{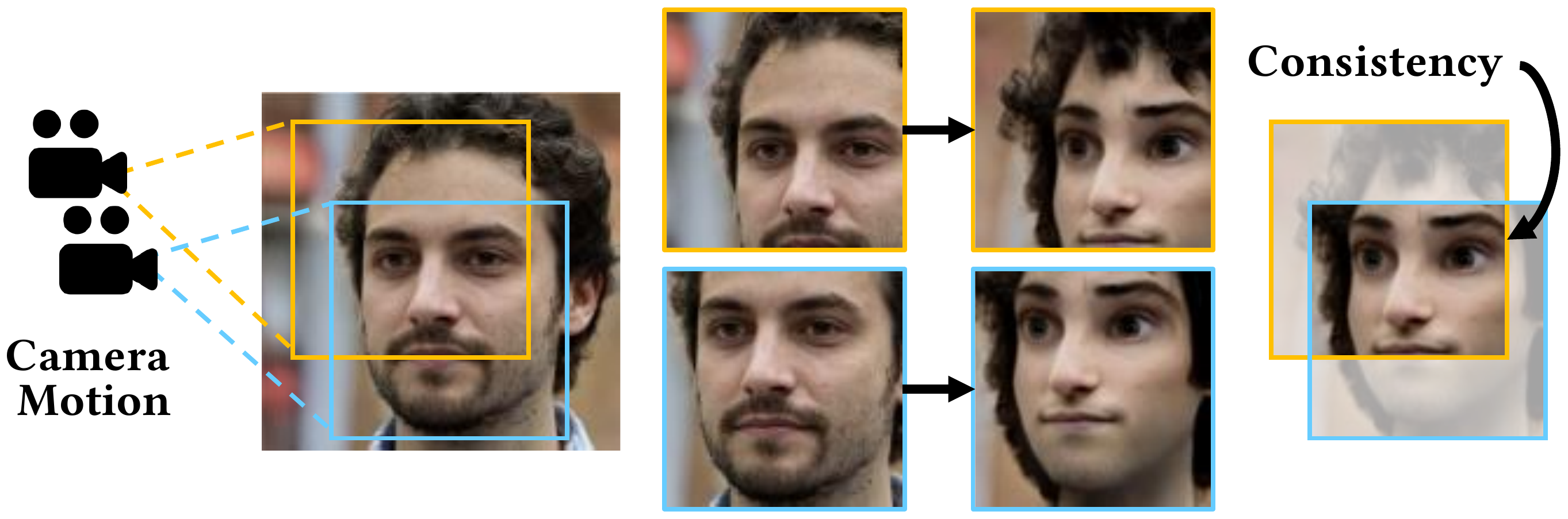}
\caption{\textbf{Illustration of the proposed flicker suppression} loss.}
\label{fig:temporal_consistency}
\end{figure}

Temporal consistency should be specially considered in our video style transfer task.
It is impractical to generate paired video data with accurate optical flows and high temporal consistency from image-oriented Toonify.
Instead, we propose to simulate the motion of camera over a single frame.
As illustrated in Fig.~\ref{fig:temporal_consistency}, the parallel movement of the camera with a limited field of view captures two subframes from a single frame. If the camera is sufficiently far away from the face, one can ignore the 3D effects and the overlapped region of these two subframes should be the same. Correspondingly, the overlapped region of their stylized results should be kept the same. This idea leads to our flicker suppression loss:
\begin{equation}
  \mathcal{L}_{\text{tmp}}=\|f_c(\widetilde{y})-G(E(f_c(x)_\downarrow), \mathbf{w}'')\|_2,
\end{equation}
where $f_c(\cdot)$ is the random cropping operation and we set one of the subframes as the whole frame for simple computation. Here, the frame $x$ is cropped in the original resolution before downsampling so that sub-pixel motion (thus temporal consistency) is supported.
Although this loss does not aim to explicitly maintain temporal coherency and does not explicitly calculate optical flow, it intrinsically assumes a uniform optical flow.
We find this simple solution effective to relieve flickers.
Our full objective takes the form of
\begin{equation}\label{eq:total_loss}
  \min_{G}\max_{D}\lambda_{\text{adv}}\mathcal{L}_{\text{adv}}+\lambda_{\text{rec}}\mathcal{L}_{\text{rec}}+\lambda_{\text{tmp}}\mathcal{L}_{\text{tmp}}.
\end{equation}

\begin{figure}[t]
\centering
\includegraphics[width=\linewidth]{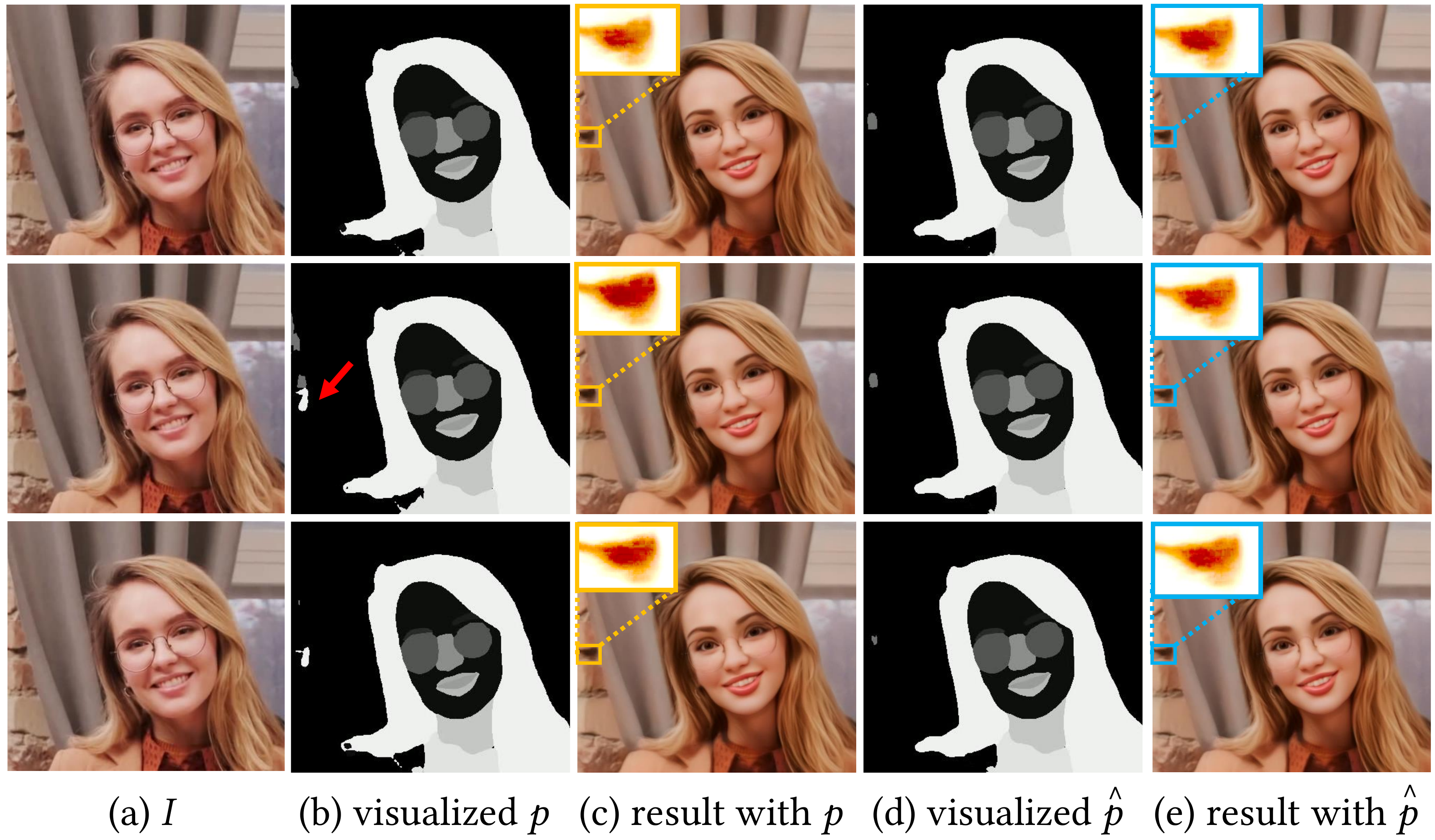}
\caption{\textbf{Effect of face parsing map smoothing.} The proposed face parsing map smoothing algorithm relieves the inconsistency between adjacent parsing maps in (b). The resulting smoothed parsing maps in (d) help synthesize more consistent results as in (e). The local regions are enlarged with their contrast enhanced for better visual comparison. In (b)(d), we visualize the parsing map by transforming it to the discrete label map. Input video: \copyright Pexels Mikael Blomkvist.}
\label{fig:smooth}
\end{figure}

\begin{figure*}[t]
\centering
\includegraphics[width=\linewidth]{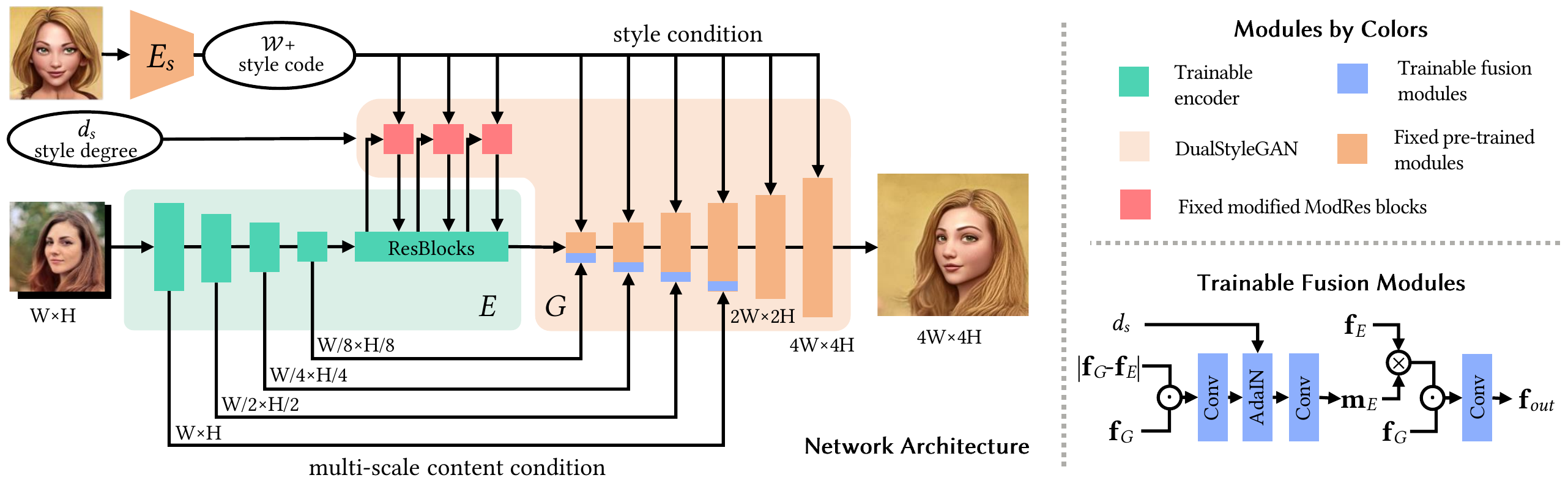}
\caption{\textbf{Framework of the proposed exemplar-based portrait video style transfer.} We use the highest 11 layers of the pre-trained DualStyleGAN and the modified ResMod blocks as our generator $G$. In addition to the extrinsic style code and the style degree $d_s$ as conditions, $G$ is also conditioned on the multi-scale content features from the encoder $E$. In the training phase, we use 256$\times$256 images as input, while during testing, our fully convolutional network accepts video frames of various sizes $W\times H$. Input video: \copyright Pexels Andrea Piacquadio.}
\label{fig:framework2}
\end{figure*}

\paragraph{Face Parsing Map Smoothing}
The temporal consistency of the stylized frames rely on the assumption that the input frames are temporally consistent, which is true for real video inputs.
However, in our framework, the video frame is augmented with its parsing map. The inconsistency in parsing maps might violate this assumption.
To solve this issue, we propose a face parsing map smoothing algorithm to strengthen the temporal consistency between consecutive face parsing maps.
First, instead of using the discrete label map as the parsing map, we use the continuous probability map before the final argmax operation of BiSeNet~\cite{Yu-ECCV-BiSeNet-2018}.
Second, we may apply an optical-flow based frame fusion to the parsing maps. Specifically, we calculate the optical flow between the frames within a temporal window. Let $I_i$ and $p_i$ denote the $i$-th frame and its parsing map, respectively, and $k$ denote the temporal window size. The operation to warp $I_j$ to $I_i$ with their optical flow $f_{j,i}$ estimated by RAFT~\cite{teed2020raft} is defined as $\theta(I_j, f_{j,i})$.
The corresponding occlusion mask is $m_{j,i}$.
Then, the parsing map $p_i$ is fused and updated with its surrounding $2k+1$ parsing maps with temporal-spatial weights:
\begin{equation}
  \hat{p}_i=\sum_{j\in[i-k,i+k]}\frac{w_{j,i}\otimes\theta(p_j, f_{j,i})}{w_i}
\end{equation}
\begin{equation}
  w_{j,i}=exp\Big(-\frac{(i-j)^2}{2\sigma_t^2}-\frac{\|I_i-\theta(I_j, f_{j,i})\|^2}{2\sigma_s^2}\Big)\otimes m_{j,i}
\end{equation}
where $w_i=\sum_{j\in[i-k,i+k]}w_{j,i}$ is the denominator for normalization and $\otimes$ is the element-wise multiplication.
Intuitively, the better matched regions in more adjacent frames have greater fusion weights.
As shown in Fig.~\ref{fig:smooth}, the flickers in the brick shadow are effectively smoothed by considering the adjacent predictions, and the brick is more consistently stylized.

\subsection{Exemplar-Based Portrait Video Style Transfer}

We move to a more complex case: exemplar-based portrait video style transfer.
This task stylizes human faces based on the style of a reference style image.
We use DualStyleGAN~\cite{yang2022Pastiche} as the backbone, which adds an extrinsic style path to StyleGAN, and is conditioned on the intrinsic style code, the extrinsic style code and the style degree.
The intrinsic style code depicts the characteristics of human faces, while the extrinsic style code characterizes the structure and color styles from the external artistic portrait image. The structure style degree $d_s$ and the color style degree $d_c$ determines the intensity of the style applied, \ie, 0 for no stylization and 1 for full stylization.
For clarity, we will use the style code to refer to the extrinsic style code in our paper.

\subsubsection{Style-Controllable Network Architecture}

The exemplar-based framework mainly follows the collection-based framework in Sec.~\ref{sec:collection}, with two modifications to realize the flexible style control as in DualStyleGAN:

\paragraph{Structure Style Control with Modified ModRes}
As in the collection-based framework, the right part of $G$ in Fig.~\ref{fig:framework2} aims to upsample the content feature and render colors and textures. Differently, the color style code $d_c$ could come either from the input frame or from an external style image for color transfer.
In DualStyleGAN, Modulative ResBlock (ModRes) is proposed to adjust the structure styles of the content features based on the style code from the external style image.
To realize such structure transfer in VToonify, we introduce ModRes into our framework (the red blocks in  Fig.~\ref{fig:framework2}) to adjust the ResBlock features of $E$. However, the original ModRes is applied to the low-resolution layers of 8$\times$8, 16$\times$16 and 32$\times$32\footnote{We empirically find that ModRes in 4$\times$4 layers of DualStyleGAN to be of little use due to the extremely low resolution. So we omit it in VToonify.}.
Its reception field and feature scale do not fully match the ResBlock features of $E$ (which are of 32$\times$32 during training).
To solve this issue, we modify the convolutional layers in each ModRes to their dilated version to re-scale the convolution kernels. Specifically, the dilation rates are $4$, $2$ and $1$ for the layers corresponding to 8$\times$8, 16$\times$16 and 32$\times$32 in DualStyeGAN, respectively.

One advantage of this modification is that we do not need to alter the pre-trained parameters of ModRes, which can be directly loaded into VToonify. Moreover, empowered by ModRes, our framework is able to support hundreds of fine-level styles in a style collection within a single model, providing users with more options.

\paragraph{Style-Degree-Aware Fusion Modules}
The structure style degree $d_s$ determines whether the stylized face matches the style better or retains more original face features as in the top right of Fig.~\ref{fig:teaser}.
The ability to flexibly adjust to which degree the original facial features are preserved is highly desired in real applications since every user has their own preferences.
Our generator built upon DualStyleGAN inherits this function, making it possible to adjust the structure degree within a single model.
Specifically, the ModRes produces residual features for structure adjustment, which are multiplied by $d_s$ and added to the ResBlock features of $E$ to generate style-degree-aware content features.

Besides the style-degree-aware content features, the mid-layer features of $E$ to characterize the original facial features are also passed to $G$.
Intuitively, these features should be extensively used for a low $d_s$ and be selectively used for a high $d_s$.
To this end, we propose a style-degree-aware fusion module as illustrated in the bottom right of Fig.~\ref{fig:framework2}.
Compared to the collection-based model, our exemplar-based model predicts an extra mask $\mathbf{m}_E$  to select the valid regions of $\mathbf{f}_E$ for fusion.
Specifically,
$\mathbf{f}_G$ and $\mathbf{f}_E$'s mismatch with $\mathbf{f}_G$ are concatenated channel-wisely and go through a convolutional layer. The result is then modulated using AdaIN conditioned by $d_s$ and fed into another convolutional layer to predict the one-channel attention mask $\mathbf{m}_E$. $\mathbf{f}_E$  is element-wisely multiplied by $\mathbf{m}_E$ and finally fused with $\mathbf{f}_G$.

\subsubsection{Data Generation and Task-Oriented Training Objectives} 
\label{sec:data_generation2}

\begin{figure}[t]
\centering
\includegraphics[width=\linewidth]{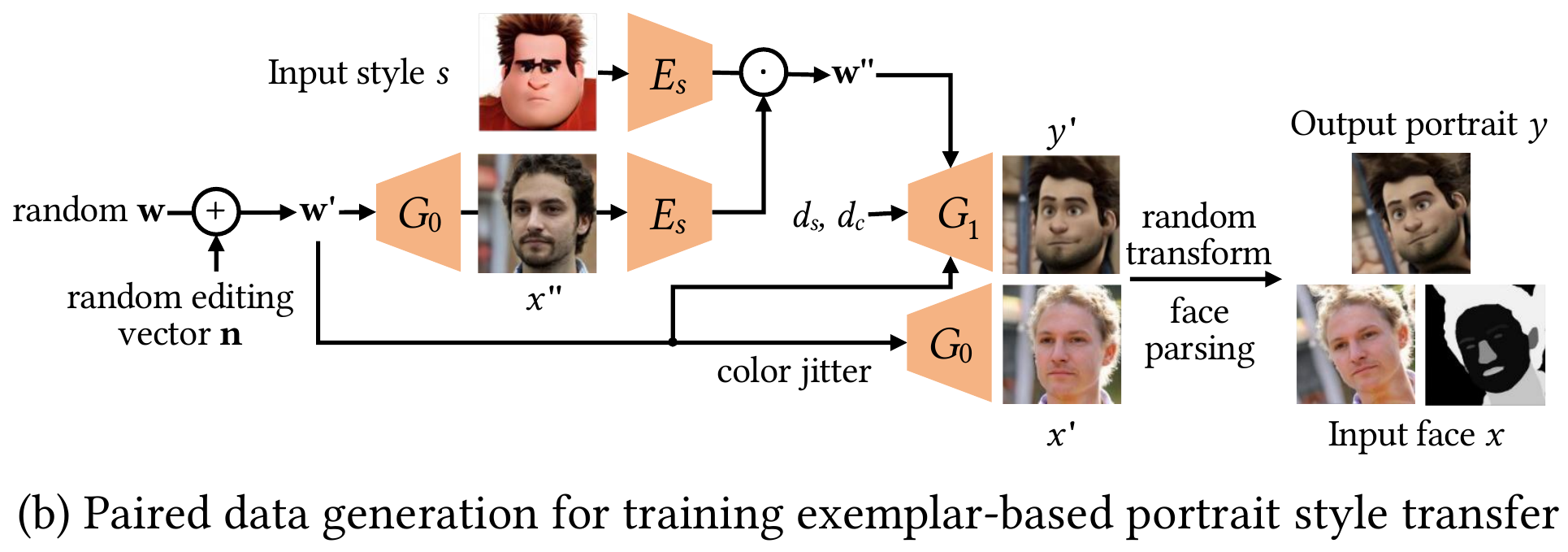}
\caption{\textbf{Pipeline of paired training data generation} for training exemplar-based portrait style transfer.}
\label{fig:data_generation2}
\end{figure}

\paragraph{Data Generation} We generate paired data based on DualStyleGAN to train our exemplar-based model. The data process is intuitively illustrated in Fig.~\ref{fig:data_generation2}. Here, $G_0$ and $G_1$ represent StyleGAN and DualStyleGAN to generate the real human face $x$ and artistic portrait $y$ in the style of the reference image $s$, respectively.
As in Sec.~\ref{sec:data_generation1}, after sampling and applying attribute editing, we obtain the style code $\mathbf{w}'$ and its corresponding face image $x''=G_0(\mathbf{w}')$. We further combine the color and texture style part of its embedded style code $E_s(x''_\downarrow)$ with the structure style part of $E_s(s)$ from the style image $s$, resulting the extrinsic style code $\mathbf{w}''$ for DualStyleGAN. Together with the intrinsic style code $\mathbf{w}'$ and the style degrees $d_s, d_c$, the stylized portrait is obtained by $y'=G_1(\mathbf{w}', \mathbf{w}'', d_s, d_c)$. To learn color transfer, we apply color jittering to $\mathbf{w}'$ by style swapping~\cite{karras2019style} (replacing the last 11 vectors of $\mathbf{w}'$ with those from another style code) and generate a face image $x'$ that is inconsistent with $y'$ in color. Finally, after geometric transformations and face parsing, we obtain the paired data $\{x,y\}$ and their style code $\mathbf{w}''$ and style degree $d_s$.

\paragraph{Model Initialization} As in Sec.~\ref{sec:data_generation1}, we initialize the fusion modules to map $\mathbf{f}_G\odot(\mathbf{f}_E\otimes\mathbf{m}_E)$  to $\mathbf{f}_G$ and pre-train $E$ to minimize
\begin{equation}
\label{eq:pretrain_loss2}
  \mathcal{L}_{\text{E}}=\|\mathbf{f}^{(\text{last})}_{E}(x''_\downarrow, \mathbf{w}'', d_s)-\mathbf{f}^{(8)}_{G_1}(\mathbf{w}', \mathbf{w}'', d_s)\|_2,
\end{equation}
where $E$ is additionally conditioned on $ \mathbf{w}''$ and $d_s$ via ModRes.
Note that the pre-training only involves the structure transfer, so the color style degree $d_c$ is omitted in Eq.~(\ref{eq:pretrain_loss2}).

\paragraph{Task-Oriented Training Objectives}
Our exemplar-based model is also driven by the reconstruction loss, the adversarial loss and the temporal consistency loss introduced in  Sec.~\ref{sec:data_generation1}. We use the class-conditional version~\cite{karras2020training} of the discriminator to accept the conditions $d_s$ and $\mathbf{w}''$. According to different  application requirements, specific training settings could be applied:
\begin{itemize}
  \item \textbf{Control structure styles.} In applications where users would like to navigate around various structure styles within a single model, we just sample different style images $s$ from the style collection to generate training data. Otherwise, using a fixed $s$ will train a model specialized in that style.
  \item \textbf{Adjust structure style degree.} In applications where users would like to navigate around results under different structure style degrees within a single model, we just sample $d_s\in[0,1]$ when generating data. We further regularize the sparsity of the attention mask $\mathbf{m}_E$ with a new loss term:
      \begin{equation}
      \label{eq:mask_loss}
          \mathcal{L}_{\text{mask}}=\lambda_{\text{mask}}\max\{\|\mathbf{m}_E\|_1/|\mathbf{m}_E|-g(d_s), 0\},
        \end{equation}
        where $|\mathbf{m}_E|$ is the element number  of $\mathbf{m}_E$ and $g(\cdot)$ is a monotonically decreasing function for $d_s\in[0,1]$. Intuitively, a large $d_s$ leads to a small $g(d_s)$, making the model predicts sparser $\mathbf{m}_E$, \ie, using less information from the input frame to give room for greater facial structure adjustment.
  \item \textbf{Control color and texture styles.} In applications where users would like to mimic the color of the reference style image, we just set $d_c=1$ when generating training data. This setting inherently supports the control of color style degree. As what $d_c$ does in DualStyleGAN, we just need to interpolate the color style codes from the style image and from the input frame during testing. Otherwise, we set $d_c=0$ and turn off color jittering (\ie, $x'$=$x''$), making the model well preserve the color and texture of the frames.
\end{itemize}
Note that the above training settings can be flexibly combined.

Obviously, training a model to support versatile stylization will inevitably sacrifice its quality.
However, we can still provide a user friendly interaction with good trade-off between the quality and flexibility.
During user interaction, the most versatile model can first provide visual previews without the need to load different checkpoints or retrain the network. After finding the ideal setting, users can then load the model specialized in that setting to obtain the high-quality, stylized output.
Considering that training on a new setting requires about one hour for VToonify, retraining a new mode is also acceptable.
Furthermore, models under commonly used settings can be pre-saved for reuse.

\section{Experimental Results}

\subsection{Implementation Details}

\paragraph{Training} The experiment is conducted on NVIDIA Tesla V100 GPUs. We use eight GPUs to train VToonify, where pre-training encoder and training full VToonify use a batch size of $1$ and $4$ per GPU, respectively. We pre-train the encoder for 30,000 iterations, taking about 50 minutes. We train VToonify for 2,000 iterations, taking about 55 minutes. Note that for each style collection, the encoder only needs to be pre-trained once. It can be shared among different training settings in Sec.~\ref{sec:data_generation2}. The paired data is generated along with the training. Thus the above reported training time includes the time for data generation.
The weights to balance different loss terms are set as
$\lambda_{\text{mse}}=0.1$,
$\lambda_{\text{perc}}=0.01$,
$\lambda_{\text{adv}}=0.01$,
$\lambda_{\text{tmp}}=1$, and
$\lambda_{\text{mask}}=0.0005$. $g(d_s)=0.1 + 0.9(1-d_s)^2$.
To smooth the face parsing map, we use $k=5$, and set $\sigma_{t}=5.5$ and $\sigma_{s}=0.2$.
We use a pre-trained pSp~\cite{richardson2020encoding} as our style encoder $E_s$.

\begin{figure*}[t]
\centering
\includegraphics[width=\linewidth]{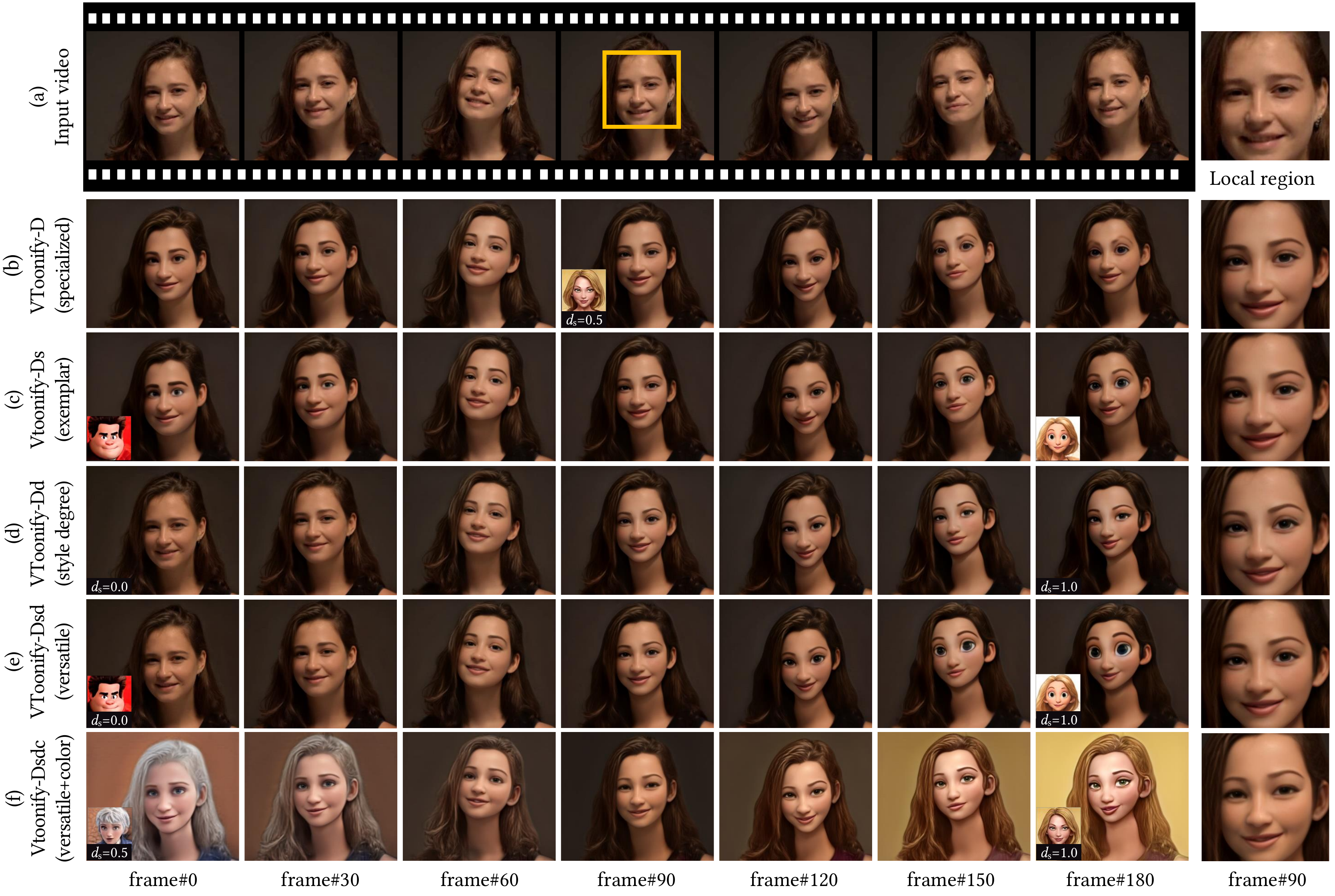}
\caption{Our method supports flexible style control on fine-level exemplar structure styles, color styles and style degree. Input video: \copyright Pexels Ketut Subiyanto.}
\label{fig:overview_style_control}
\end{figure*}

Since our method has several training settings for different tasks, we name each model individually, as summarized in Table~\ref{tb:name} for clarify. For example,  VToonify-D without suffix means a model built upon DualStyleGAN specialized for one exemplar style, while
VToonify-Dsd indicates a versatile model supporting both structure style control and style degree adjustment.

\paragraph{Testing} For each video, we follow \cite{karras2019style} to detect, align the face in the first frame and crop 256$\times$256 images for pSp to obtain the style code. 
This style code is used for all other frames. We apply BiSeNet~\cite{Yu-ECCV-BiSeNet-2018} to extract face parsing map for each frame. We find the original parsing maps work well for most of the videos. Thus by default, we do not apply the face parsing map smoothing algorithm in the experiment. It is optional for videos with inaccurately predicted parsing maps.
We use the FaceForensics++~\cite{roessler2019faceforensicspp} and videos from Pexels as our testing dataset.
We proprocess the original videos by detecting the face in the first frame, and resizing the frame so that the eye distance is 64 pixels. Centered on the eyes, we crop the first frame to almost 400$\times$400. All other frames use the same resizing and cropping parameters as the first frame to generate the final video for testing.

\begin{table}[b]
\begin{center}
\newcommand{\tabincell}[2]{\begin{tabular}{@{}#1@{}}#2\end{tabular}}
\caption{\textbf{Summary of VToonify models and settings}.}
\begin{tabular}{c|l}
\toprule
\makecell[c]{\textbf{Name} \\ \textbf{(Backbone)}}   & \textbf{Settings} \\
\midrule
\makecell[c]{VToonify-T \\ (Toonify)}  &collection-based model \\
\midrule
 &  exemplar-based model, with suffix $*$ of: \\
VToonify-D$*$ & `s': one model for various structure styles \\
(DualStyleGAN) & `d': one model for various style degrees \\
 & `c': one model for various color styles \\
\bottomrule
\end{tabular}
\label{tb:name}
\end{center}
\end{table}

\subsection{Flexible Style Control}
\label{sec:ex_style_control}

This section elaborates the flexibility of VToonify in style control.
We compare the performance of our model trained on different tasks on the same video clip, as shown in Fig.~\ref{fig:overview_style_control}.
In Fig.~\ref{fig:overview_style_control}(b), VToonify-D specific to $d_s=0.5$ and a fixed style synthesizes high-quality frames.
Figures~\ref{fig:overview_style_control}(c)-(f) show the results of VToonify-Ds, VToonify-Dd, VToonify-Dsd and VToonify-Dsdc, respectively.
The conditional inputs of $d_s$ and the reference style image are shown in the lower left of frame\#0 and frame\#180.
All models use the same condition as VToonify-D in frame\#90 for quality comparison.
It can be seen that VToonify-Ds supports fine-level control by choosing exemplar style images, and the facial structure style gradually becomes more adorable in Fig.~\ref{fig:overview_style_control}(c).
VToonify-Dd supports adjusting the style degree. With $d_s=0$, our model purely super-resolves the input frame, while $d_s=1$ leads to strong styles. Here, setting $d_s=0.5$ provides a balance in facial feature preservation and artistry.
VToonify-Dsd is a versatile model to support both of the above controls.
From the local regions, it can be seen that although the capacity of these models differs, they demonstrate similar performance, possibly due to the good distillation of the backbones. The main difference is that the specialized VToonify-D can better preserves the details.

The main sacrifice of the quality comes from color style control, as in Fig.~\ref{fig:overview_style_control}(f). In this case, since the training pairs have distinct color and texture discrepancies, the model has to rely more on the style condition than the content condition. As a result, VToonify-Dsdc cannot preserve the details, such as the double eyelid. However, the facial region is general still of high quality as other models.

\begin{figure*}[t]
\centering
\includegraphics[width=0.98\linewidth]{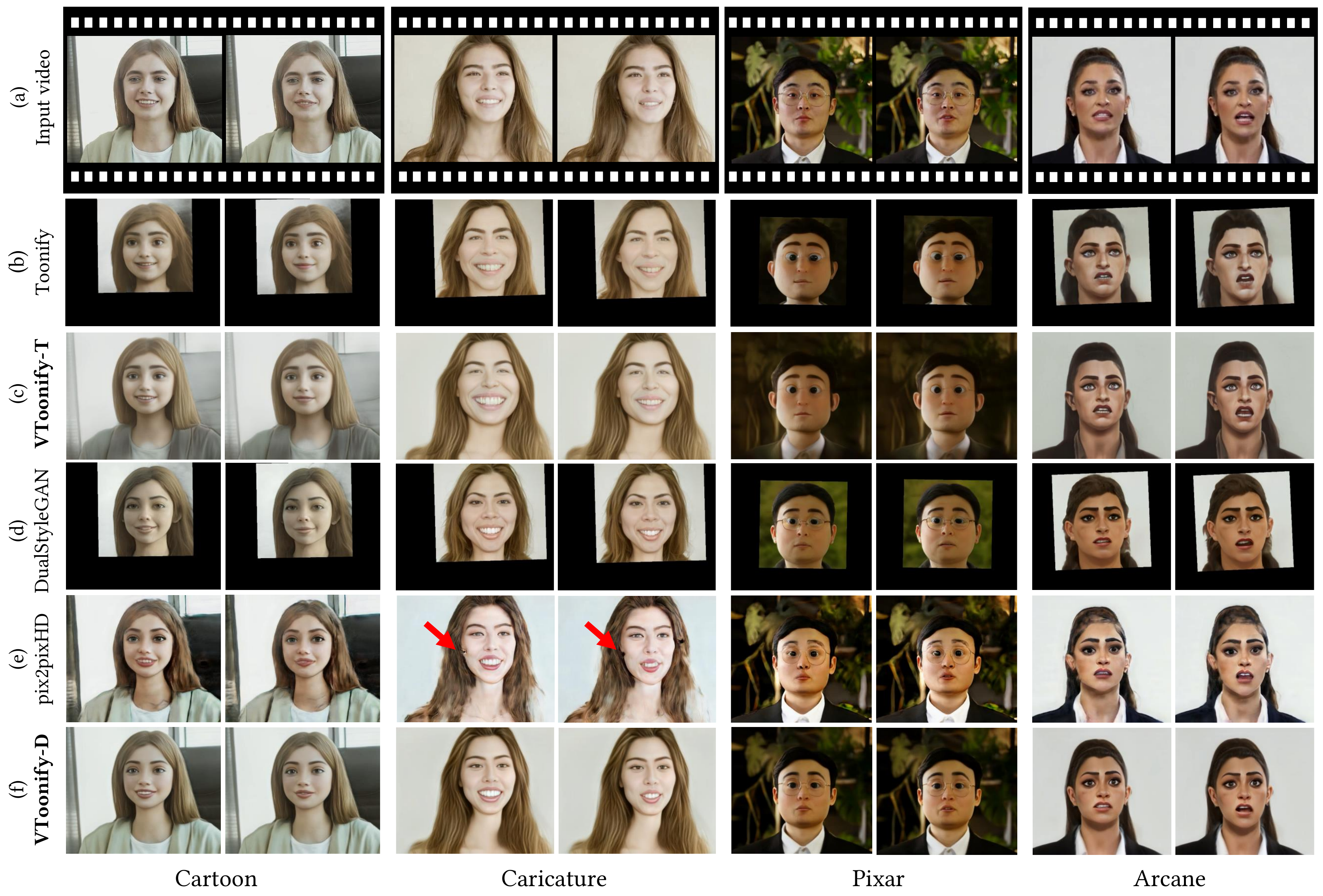}
\caption{\textbf{Visual comparison with face style transfer methods on four style collections}. Input video: \copyright Pexels Tima Miroshnichenko, Kampus Production, Anthony Shkraba, Rodnae Productions.}
\label{fig:compare1}
\end{figure*}

\subsection{Comparison with State-of-the-Art Methods}

\paragraph{Comparison Methods} Since there are few methods that exactly handle our task, we compare with the most related ones and adapt them to our task.
Besides our two backbones, we choose a high-resolution image-to-image translation baseline for StyleGAN distillation, and two image animation baselines for exemplar-based video generation. Specifically,
\begin{itemize}
  \item \textbf{Toonify}~\cite{pinkney2020resolution} is a StyleGAN-based backbone for collection-based style transfer on aligned faces. We follow \cite{karras2019style} to align faces and crop 256$\times$256 images for pSp to obtain the style codes. The style codes are fed into Toonify to obtain 1024$\times$1024 stylized result. Finally we re-align the result back to its original position in the video. The unstylized region is simply set to black.
  \item \textbf{DualStyleGAN}~\cite{yang2022Pastiche} is a StyleGAN-based backbone for exemplar-based style transfer. We apply the same data preprocessing and postprocessing as in Toonify.
  \item \textbf{Pix2pixHD}~\cite{Wang2017High} is an image-to-image translation model, which is widely used to distill pre-trained models for high resolution editing~\cite{viazovetskyi2020stylegan2}. We train it on our paired data. We follow pix2pixHD to use our extracted parsing map as its additional instance map inputs.
  \item \textbf{First Order Motion (FOM)}~\cite{siarohin2019first} is a representative image animation model. It is trained on 256$\times$256 images. and has degraded performance using other image sizes. Therefore, we first resize the video frames to 256$\times$256 for FOM to animate and resize the results to their original size. For a fair comparison, FOM uses the first stylized frame of our method as its reference style image.
  \item \textbf{DaGAN}~\cite{hong2022depth} is an image animation model considering 3D face information. We apply the same data preprocessing and postprocessing as in FOM.
\end{itemize}

\begin{figure}[t]
\centering
\includegraphics[width=\linewidth]{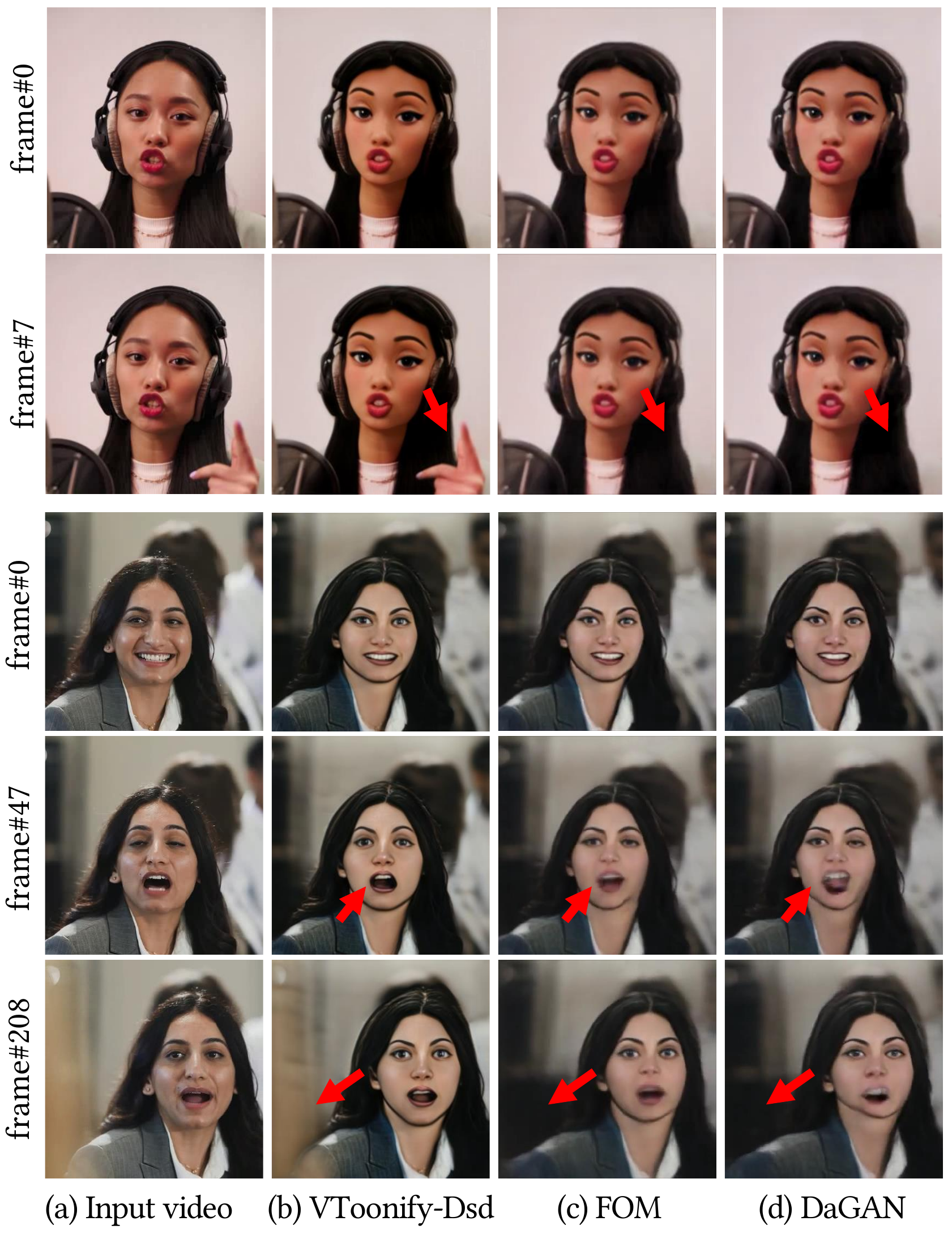}
\caption{\textbf{Visual comparison with image animation methods}. Input video: \copyright Pexels Anthony Shkraba and Rodnae Productions.}
\label{fig:compare2}
\end{figure}

\begin{table}[htbp]
\begin{center}
\newcommand{\tabincell}[2]{\begin{tabular}{@{}#1@{}}#2\end{tabular}}
\caption{\textbf{User preference rates} in terms of style similarity, content preservation, temporal consistency and overall quality.}
\begin{tabular}{l|cccc}
\toprule
\textbf{Method}  & \textbf{Style} & \textbf{Content} & \textbf{Temporal} & \textbf{Overall}\\
\midrule
Toonify & 7.4\% & 5.5\% & 4.5\% & 3.9\% \\
Vtooniy-T & 31.3\% & 21.6\% & 29.4\% & 31.9\%\\
DualStyleGAN & 1.0\% & 1.0\% & 0.0\% & 0.3\% \\
pix2pixHD & 11.9\% & 9.0\% & 5.5\% & 9.0\% \\
VToonify-D & \textbf{48.4\%} & \textbf{62.9\%} & \textbf{60.6\%} & \textbf{54.9\%}\\
\midrule
FOM & 8.4\% & 5.8\% & 6.8\% & 4.5\% \\
DaGAN & 1.6\% & 3.6\% & 2.2\% & 1.6\%\\
VToonify-Dsd & \textbf{90.0\%} & \textbf{90.6\%} & \textbf{91.0\%} & \textbf{93.9\%} \\
\bottomrule
\end{tabular}
\label{tb:user_study}
\end{center}
\end{table}

\paragraph{Comparison with Face Style Transfer Methods}
Figure~\ref{fig:compare1} shows the qualitative comparison with three high-resolution models: Toonify, DualStyleGAN and pix2pixHD.
Since pix2pixHD is trained on paired data generated by DualStyleGAN in a specific style setting, we correspondingly use the specialized VToonify-D for comparison.
It can be seen that VToonify-T and VToonify-D surpass their corresponding backbones Toonify and DualStyleGAN in stylizing the complete video, while maintaining the same high quality and visual features as the backbones for each single frame. For example, VToonify-T follows Toonify to impose a strong style effect like round eyes in the Pixar style. VToonify-D, on the other hand, is better at preserving the facial features.
Compared with VToonify-D, pix2pixHD suffers from flickers and artifacts. For example, in the Caricature style, there are black holes near the cheek in pix2pixHD's results.

Since pix2pixHD and our method are not trained on real data, it is not feasible to use the common metrics like IS and FID.
Therefore, we report user preference scores for quantitative evaluations. We conduct a user study, where 31 subjects are invited to select what they consider to be the best results from the five style transfer methods in terms of 1) how well the result matches the style of the style collection, 2) how well the result preserves the details of the original video, 3) the temporal consistency of the result and 4) the overall quality of video toonification. Ten testing videos from FaceForensics++ are used.  Table~\ref{tb:user_study} summarizes the average preference rates, where VToonify-D receives the best rates in all four metrics.

\paragraph{Comparison with Image Animation Methods} We present a visual comparison with FOM and DaGAN in Fig.~\ref{fig:compare2}.
Since image animation frameworks can handle arbitrary style per model, we correspondingly use the versatile VToonify-Dsd for comparison.
The first video has one challenge: new content.
The image animation frameworks inherently cannot create information beyond its reference style image, \eg, hands appearing in frame\#7.
By comparison, VToonify-Dsd is robust to the new content.

The second video has large motions.
Although image animation methods successfully animate the stylized frame\#0 based on the motions in the video,
when the face expression changes drastically, FOM and DaGAN fail to generate the mouth details, leading to a blurry face.
By comparison, the results of our method show rich facial details consistently throughout the video.

To quantitatively evaluate VToonify-Dsd, we perform a user study in the same setting as VToonify-T and VToonify-D, where
31 subjects are invited to evaluate the results on ten testing videos from FaceForensics++.
Table~\ref{tb:user_study} reports the average preference rates, where VToonify-Dsd receives the best rates in all four metrics.

\paragraph{Running Time}
Table~\ref{tb:running_time} reports the running time
(excluding video reading/writing) of the compared methods on a NVIDIA Tesla V100 GPU with a batch size of $1$.
Pix2pixHD is faster than the three frameworks based on the big StyleGAN model.
The running time of VToonify is similar to Toonify and DualStyleGAN on aligned 1024$\times$1024 videos.
For unaligned 1600$\times$1280 videos, Toonify and DualStyleGAN need to detect and crop faces every frame, making them less efficient than VToonify. VToonify surpasses FOM and DaGAN for generating higher-resolution videos in comparable time.
Since our method handles every frame independently, it can be further accelerated by parallel processing.

\begin{table}[]
\begin{center}
\newcommand{\tabincell}[2]{\begin{tabular}{@{}#1@{}}#2\end{tabular}}
\caption{\textbf{Running time} on different video size (second per frame).}
\begin{tabular}{l|ccc}
\toprule
\textbf{Output Size}  & 1024$\times$1024 & 1600$\times$1280  & 256$\times$256\\
\midrule
Toonify & 0.117 & 0.389 & - \\
DualStyleGAN & 0.128 & 0.402 & - \\
pix2pixHD & 0.015 & 0.017 & - \\
FOM & - & - & 0.169 \\
DaGAN & - & - & 0.059 \\
VToonify & 0.106 & 0.198 & - \\
\bottomrule
\end{tabular}
\label{tb:running_time}
\end{center}
\end{table}

\subsection{Ablation Study}

\begin{figure}[t]
\centering
\includegraphics[width=\linewidth]{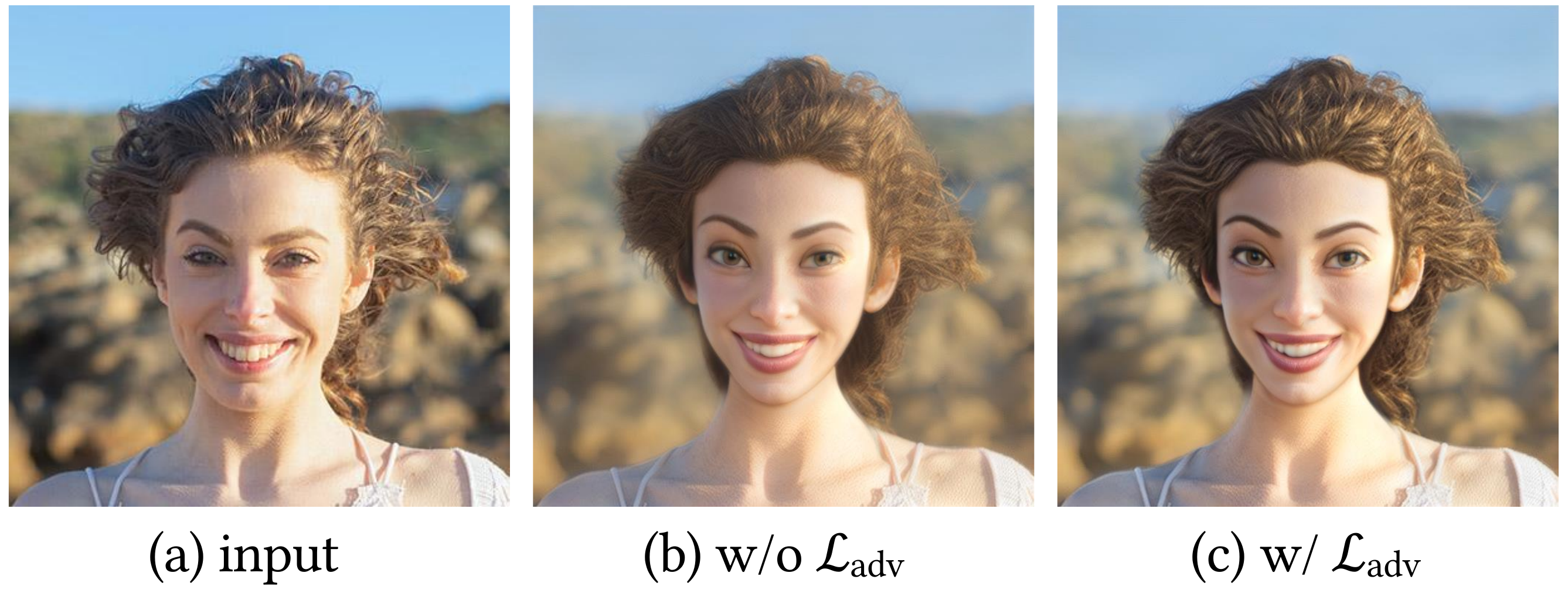}
\caption{\textbf{Effect of $\mathcal{L}_{\text{adv}}$} in reinforcing the realism of the output. Input video: \copyright Pexels Kampus Production.}
\label{fig:adv}
\end{figure}

\begin{figure}[t]
\centering
\includegraphics[width=\linewidth]{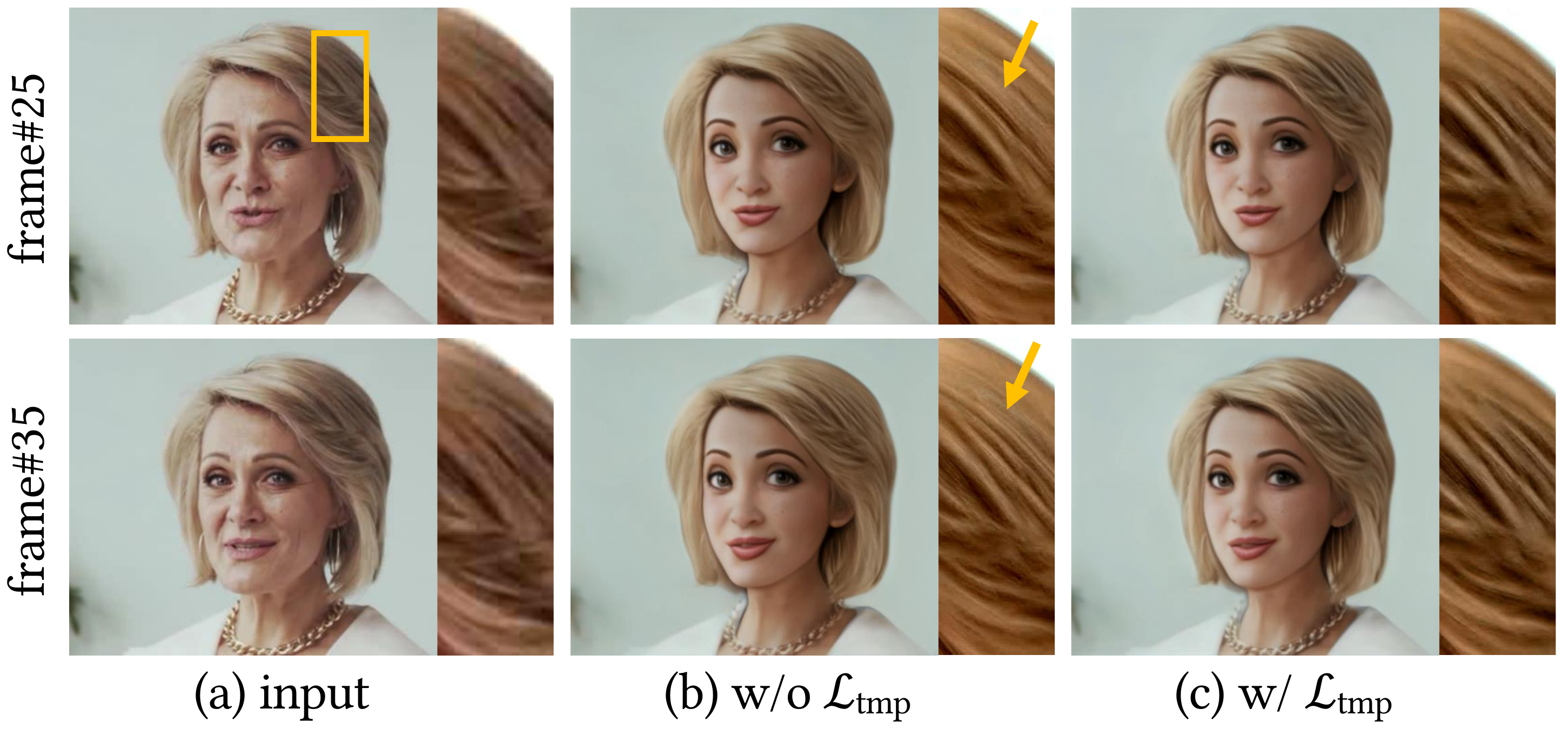}
\caption{\textbf{Effect of $\mathcal{L}_{\text{tmp}}$} in achieving temporal consistency. The local regions are enlarged with their contrast enhanced for better visual comparison. Input video: \copyright Pexels Anthony Shkraba.}
\label{fig:temporal_loss}
\end{figure}

\begin{figure}[t]
\centering
\includegraphics[width=\linewidth]{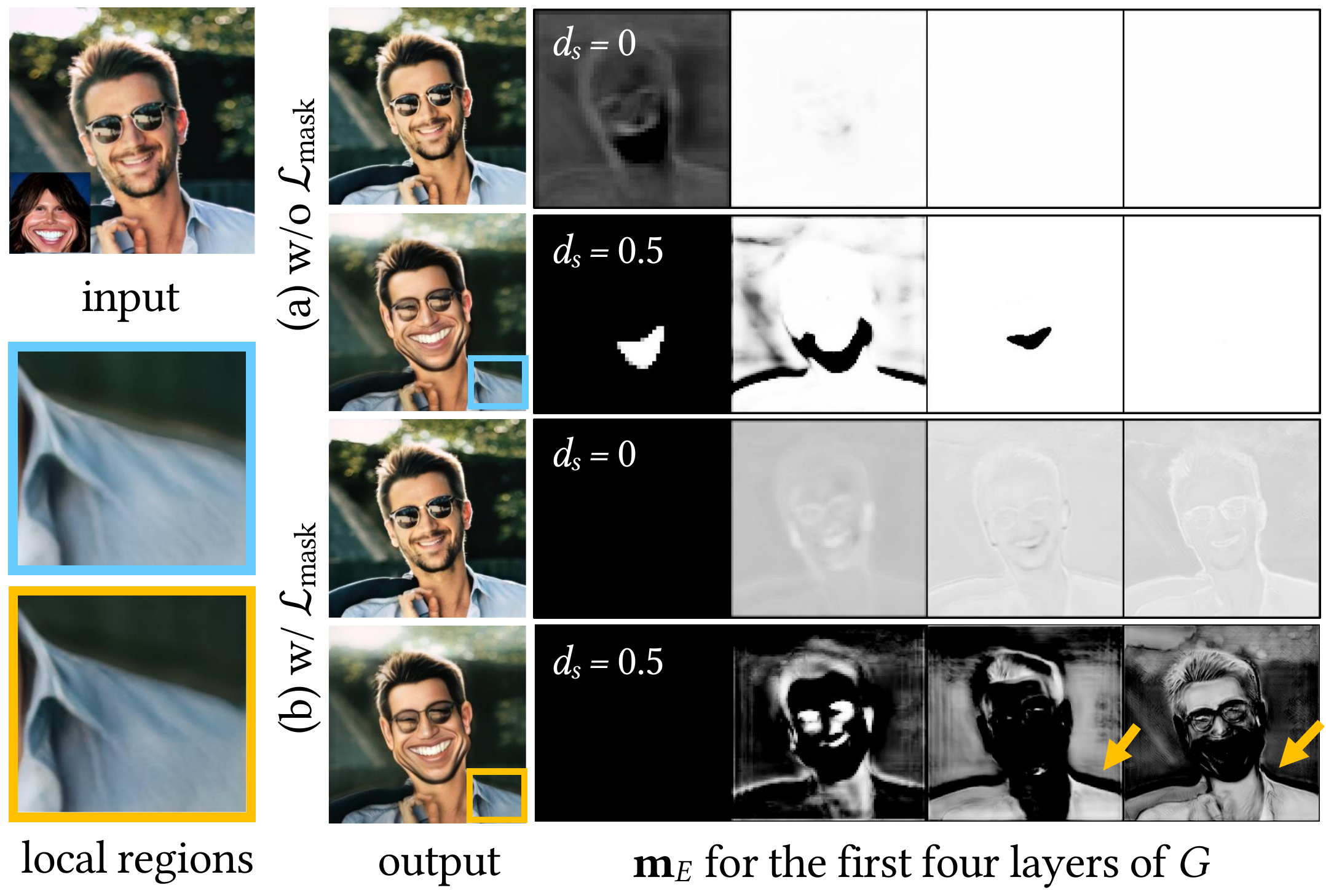}
\caption{\textbf{Effect of $\mathcal{L}_{\text{mask}}$} in eliminate ghosting artifacts. Input video: \copyright Pexels Chloe.}
\label{fig:mask_loss}
\end{figure}

\noindent
\paragraph{Loss Functions} Figure~\ref{fig:adv} compares the results with and without the adversarial loss $\mathcal{L}_{\text{adv}}$ in Eq.~(\ref{eq:total_loss}).
Without $\mathcal{L}_{\text{adv}}$, the results have low contrast and blurry hair details. The adversarial training effectively enriches the texture details,
enhancing the overall realism.
The effect of the temporal consistency term in Eq.~(\ref{eq:total_loss}) is shown in Fig.~\ref{fig:temporal_loss}.
Without $\mathcal{L}_{\text{tmp}}$, the same strand of hair is stylized into different textures in two consecutive frames, leading to annoying flickers.
The proposed temporal consistency loss, though simple, solves this issue. 
In Fig.~\ref{fig:mask_loss}, we study the effect of the mask loss $\mathcal{L}_{\text{mask}}$ in Eq.~(\ref{eq:mask_loss}).
We compare the results with and without $\mathcal{L}_{\text{mask}}$, and show the estimated masks $\mathbf{m}_E$ in four fusion modules in $G$.
The mask loss is proposed for style degree adjustment. For $d_s=0$, both settings effectively leverage the mid-layer content features to reconstruct the input frame. However, when $d_s=0.5$, the model without $\mathcal{L}_{\text{mask}}$ overuses the content features, leading to the ghosting artifacts at the edge of clothes. By comparison, the mask loss imposes the sparsity of $\mathbf{m}_E$, forcing the model to select only the valid information, \eg, regions inside the clothes and forehead, avoiding the ghosting artifacts. Note that the edge of clothes has low values in $\mathbf{m}_E$ as pointed by the yellow arrows.

\begin{figure}[t]
\centering
\includegraphics[width=\linewidth]{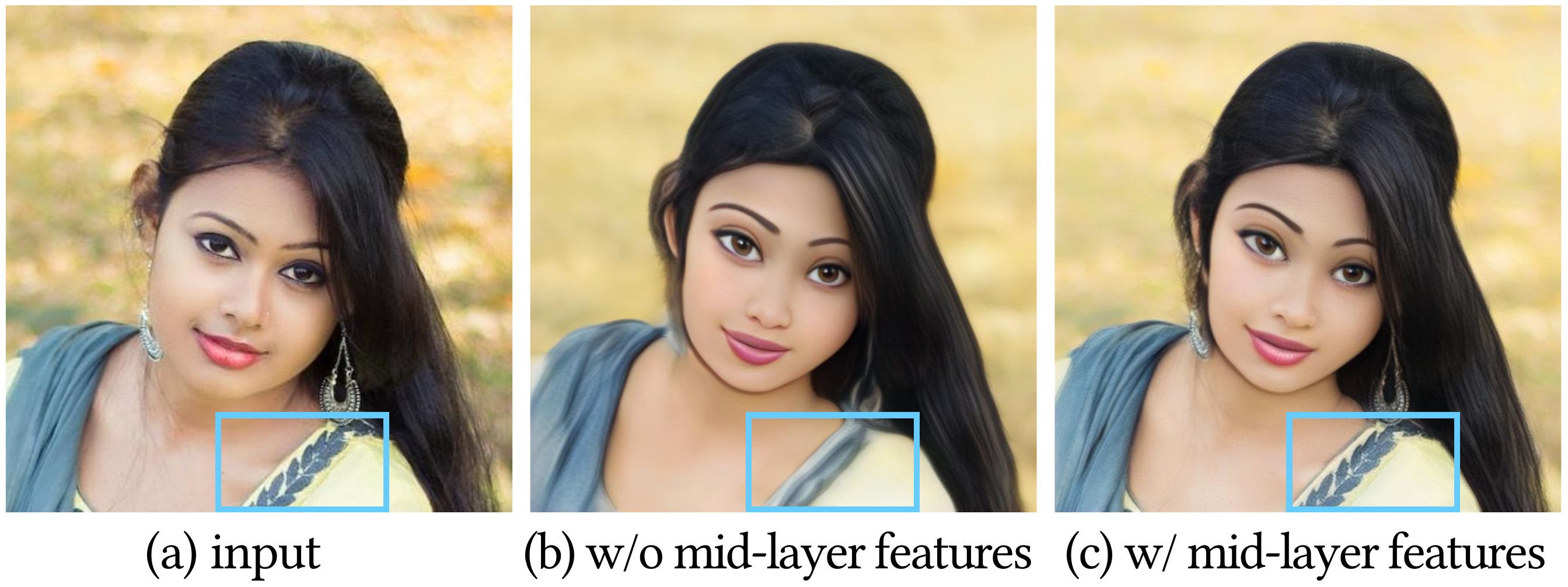}
\caption{\textbf{Effect of multi-scale content features} in preserving the visual details of the input frame. Input video: \copyright Pexels Joy Deb.}
\label{fig:skip}
\end{figure}

\paragraph{Multi-Scale Content Features} In addition to $\mathcal{L}_{\text{mask}}$, we further investigate the multi-scale content features in preserving the details.
We train a baseline model by manually setting all $\mathbf{m}_E=\mathbf{0}$ to remove the mid-layer content features. As shown in Fig.~\ref{fig:skip}(b), with only a single layer content feature, the hair and the patterns in the clothes becomes blurry. By using multi-scale content features, these important details, which are hard to encode and reconstruct within DualStyleGAN, are effectively synthesized.

\begin{figure}[t]
\centering
\includegraphics[width=\linewidth]{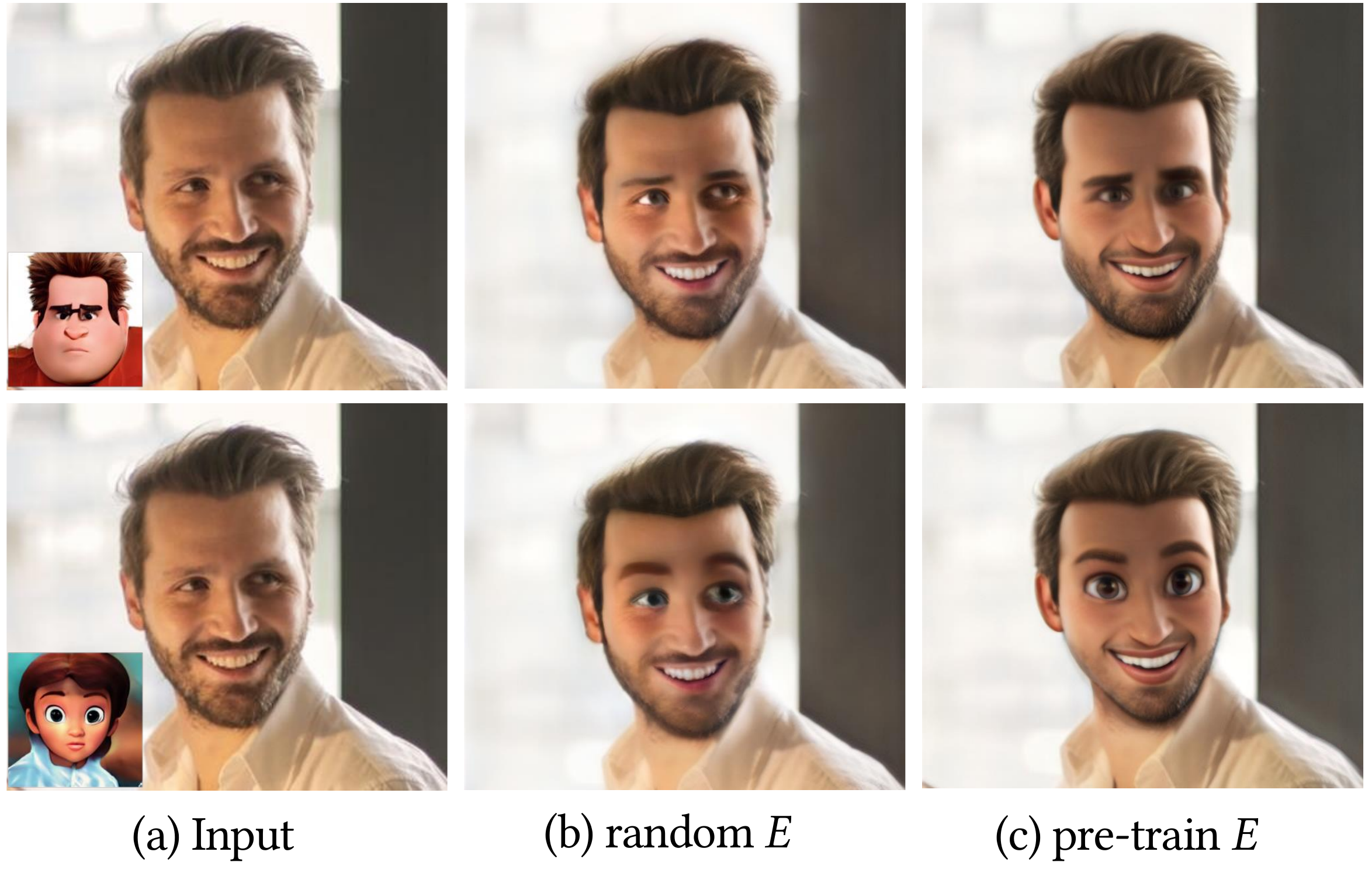}
\caption{\textbf{Effect of pre-training the encoder} in model distillation. Input video: \copyright Pexels Andrea Piacquadio.}
\label{fig:ablation_pretrain}
\end{figure}

\paragraph{Encoder Pre-Training} Figure~\ref{fig:ablation_pretrain} compares VToonify-Dsd trained with randomly initialized $E$ and with pre-trained $E$.
Without pre-training, our model fails to capture the structure style of the reference style image. The pre-training strategy makes it easier to distill DualStyleGAN for effective exemplar-based style transfer.

\begin{figure}[t]
\centering
\includegraphics[width=\linewidth]{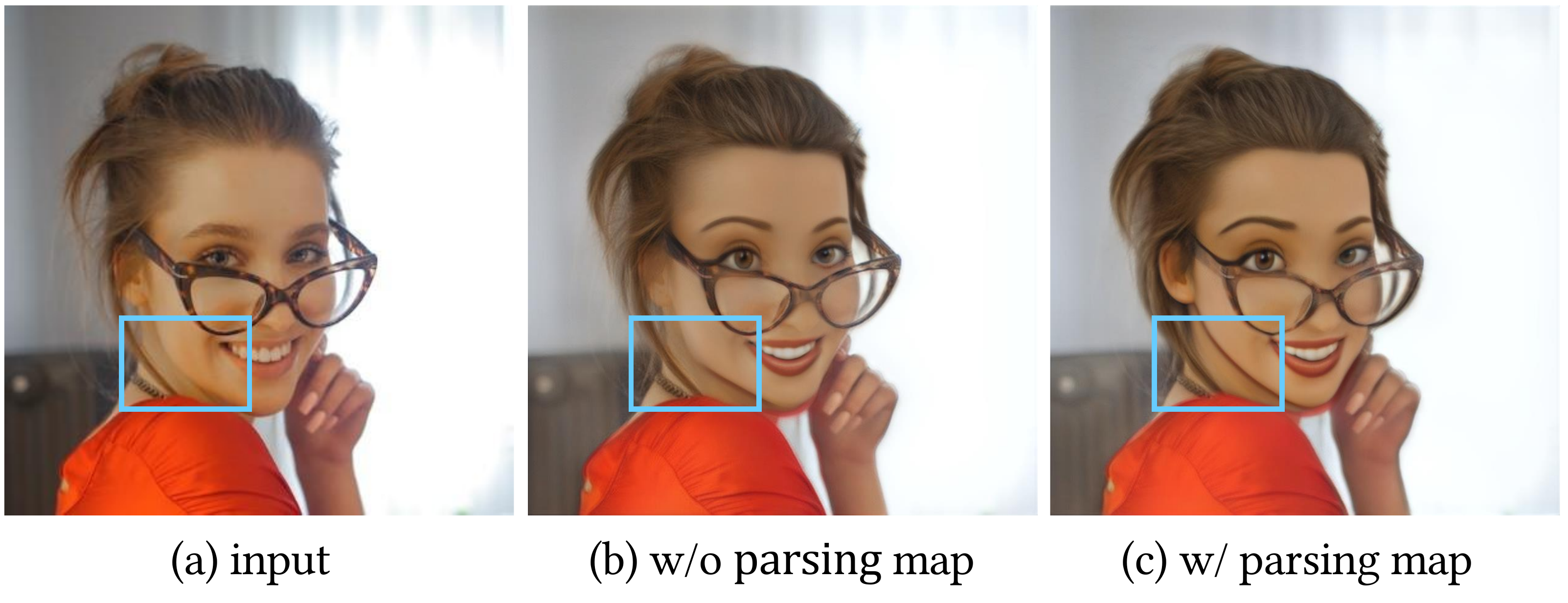}
\caption{\textbf{Effect of face parsing maps} in generating cartoon-like face structures. Input video: \copyright Pexels Andrea Piacquadio.}
\label{fig:ablation_parsing}
\end{figure}

\paragraph{Parsing map} Figure~\ref{fig:ablation_parsing} compares VToonify-Dsd trained with and without face parsing maps as additional inputs. It can be seen that the face parsing map helps our model to generate more accurate structures of the face.

\subsection{Limitations}

\begin{figure}[t]
\centering
\includegraphics[width=\linewidth]{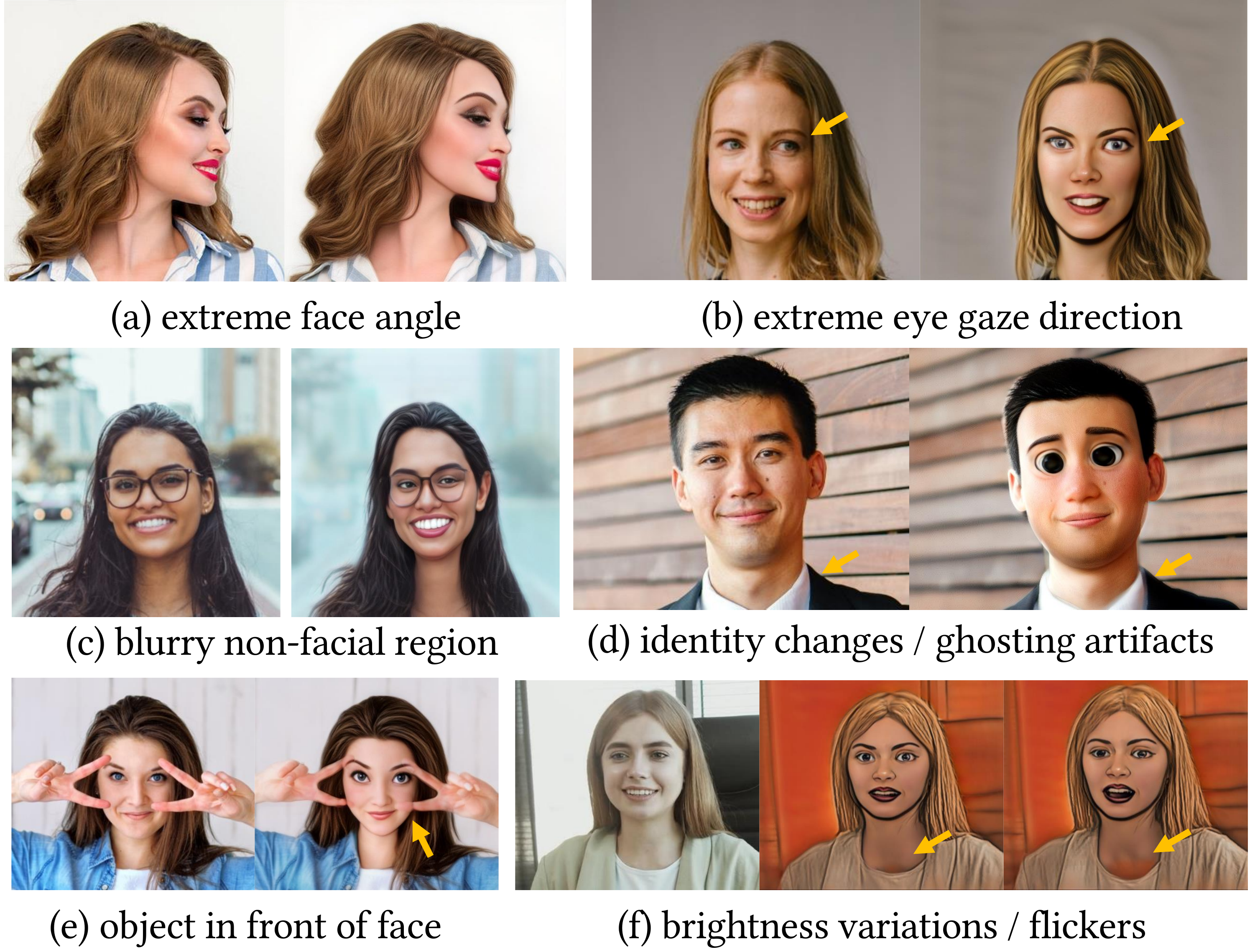}
\caption{\textbf{Failure cases of VToonify}. Input video: \copyright Pexels Anastasiya Gepp, Mentatdgt, Daniel Xavier, Pixabay, Tima Miroshnichenko.}
\label{fig:limitation}
\end{figure}

In this section, we discuss limitations of our proposed method.
\begin{enumerate}
\item \textbf{Data/model bias}. Our framework distills both the data and the model from the StyleGAN-based backbones, thus suffering the data bias and model bias. For example, as with most StyleGAN-based models, our method cannot handle faces in extreme angles (unable to adjust the facial structure in Fig.~\ref{fig:limitation}(a)) or with extreme eye gaze direction (wrong direction in Fig.~\ref{fig:limitation}(b)), either.
Another example is that DualStyleGAN tends to synthesize images with blurry backgrounds under the Caricarure style. Thus our method cannot preserve the details of non-facial regions in this style as shown in Fig.~\ref{fig:limitation}(c).
The bias could also lead to failure of reference style preservation. For example, as reported in DualStyleGAN, uncommon styles like the extremely large cartoon eyes cannot be well imitated. The style is sometimes not respected with the original color styles. For example, the results in Fig.~\ref{fig:overview_style_control}(b)-(e) look photo-realistic, less similar to the reference cartoon style.
It is difficult to overcome these limitations inherited from our backbones. But in turn, our framework benefits from stronger backbones, which is promising with the fast advancement in computer graphics and vision.
\item  \textbf{Identity Preservation}. Unlike existing models for caricature generation~\cite{cao2018carigans}, which aims to achieve artistry and identity preservation at the same time, in our task, we aim to change facial features such as the round cartoon eyes (Fig.~\ref{fig:limitation}(d)) following our backbones. Therefore, the identity might be not well preserved with a large $d_s$.
\item  \textbf{Ghosting artifacts}. The artifacts mainly come from the scale mismatches between the stylized facial region and remaining regions. For example, Pixar portrait has a large head and a thin neck. Therefore, there is a gap between the shrinked neck and the collar, leading to ghosting artifacts as in Fig.~\ref{fig:limitation}(d).
\item \textbf{Occlusions}. Our method is less effective in handling objects in the facial region, \eg, hands in Fig.~\ref{fig:limitation}(e). Unlike the non-facial region, our method relies less on the mid-layer content features in the facial regions as visualized in Fig.~\ref{fig:mask_loss}(b). One possible solution is to manually overlap random occlusions on the training images. Another similar problem is the missing of the small earrings in Fig.~\ref{fig:limitation}(c).
\item \textbf{Color control with artifacts}. As pointed out in Sec.~\ref{sec:ex_style_control}, enabling color control forces the model to rely less on the mid-layer content features, which not only losses the details but also enlarges the small flickers in the original video. We observe ignorable variations in saturated regions sometime lead to noticeable brightness variations in Fig.~\ref{fig:limitation}(f). In future work, we would like to explore a better way of color control, \eg, combining our versatile model VToonify-Dsd with state-of-the-art color transfer methods.
\item \textbf{Flickers}. Figure~\ref{fig:limitation}(f) also shows the flickers in the background. Our flicker suppression loss is based on a simple uniform optical flow, thus might be less effective on scenes with saturated regions or complex motions.
\end{enumerate}

\section{Conclusion}

In this paper, we propose a new VToonify framework for style controllable high-resolution video toonification.
By distilling StyleGAN-based image toonification models in terms of both their synthetic data and network architectures,
our framework presents high performance in handling videos and provides flexible control over the structural style, color style and style degree.
While existing StyleGAN-based image editing methods focus on encoding faces into the style latent space for processing,
we show that combining the high-level style code and multi-level content features with spatial resolution can better reconstruct the image details, especially for non-facial objects. It also overcomes StyleGAN's inherent limitation of fixed resolution and aligned faces.
We believe the idea of our framework design is not limited to face style transfer, which can be potentially applied to other image and video editing tasks such as image super-resolution and facial attribute editing.

\begin{acks}
The authors would like to thank Yuming Jiang for proofreading.
This study is supported under the RIE2020 Industry Alignment Fund Industry Collaboration Projects (IAF-ICP) Funding Initiative, as well as cash and in-kind contribution from the industry partner(s).
\end{acks}

\bibliographystyle{ACM-Reference-Format}
\bibliography{sample-bibliography}


\end{document}